\documentclass{article} 
\usepackage{preprint,times}

\usepackage[utf8]{inputenc} 
\usepackage[T1]{fontenc}    
\usepackage{amsmath}
\usepackage{amsfonts}       
\usepackage{graphicx}
\usepackage{booktabs}       
\usepackage{multirow}
\usepackage{wrapfig}
\usepackage{verbatim}
\usepackage{nicefrac}       
\usepackage{microtype}      
\usepackage[dvipsnames,table]{xcolor}
\usepackage{pifont}         
\usepackage{caption}
\usepackage{xspace}
\usepackage{hyperref}       
\usepackage{url}            

\definecolor{myblue}{RGB}{245,245,250}

\newcommand{\name}{InterTab\xspace}
\newcommand{\dataset}{\textbf{InterTab-22K}\xspace}

\newcommand{\eg}{\emph{e.g., }\xspace}
\newcommand{\ie}{\emph{i.e., }\xspace}

\title{\name: Interleaved Visual-Structure Alignment for Multimodal Table Reasoning}

\makeatletter
\newcommand{\authornote}[2]{%
  \protected@xdef\@thanks{\@thanks\protect\footnotetext[#1]{#2}}%
}
\makeatother
\author{%
\textbf{Hanqian Li}\textsuperscript{\rm 1,4*\ddag},
\textbf{Sirui Huang}\textsuperscript{\rm 2*\dag},
\textbf{Chen Ling}\textsuperscript{\rm 3,4*\ddag},
\textbf{Jungang Li}\textsuperscript{\rm 1,5},
\textbf{Yu Huang}\textsuperscript{\rm 6}, \\
\textbf{Kening Zheng}\textsuperscript{\rm 7},
\textbf{Yonghua Hei}\textsuperscript{\rm 1,5},
\textbf{Xiangrong He}\textsuperscript{\rm 3},
\textbf{Shiyi Wang}\textsuperscript{\rm 4},
\textbf{Pengcheng Zhu}\textsuperscript{\rm 4}, \\
\textbf{Dongnan Liu}\textsuperscript{\rm 4},
\textbf{Wei Zhou}\textsuperscript{\rm 4},
\textbf{Linjian Mo}\textsuperscript{\rm 4},
\textbf{Nai Ding}\textsuperscript{\rm 3},
\textbf{Xuming Hu}\textsuperscript{\rm 1,5\dag} \\
\textsuperscript{\rm 1}{The Hong Kong University of Science and Technology (Guangzhou)} \\
\textsuperscript{\rm 2}{Hong Kong Polytechnic University}
\textsuperscript{\rm 3}{Zhejiang University}
\textsuperscript{\rm 4}{Ant Group} \\
\textsuperscript{\rm 5}{The Hong Kong University of Science and Technology} \\
\textsuperscript{\rm 6}{City University of Hong Kong}
\textsuperscript{\rm 7}{University of Illinois at Chicago}
\authornote{1}{Equal contribution.}%
\authornote{2}{Corresponding author: sirui.huang@connect.polyu.hk, xuminghu@hkust-gz.edu.cn.}%
\authornote{3}{Work done during an internship at Ant Group.}
}

\begin{document}

\maketitle

\begin{abstract}
Table images preserve structural information that are often lost in text serialization, and reasoning over them requires locating relevant rows, columns, and cells step by step. Current multimodal large language models (MLLMs) encode the whole image once before reasoning, so they cannot pick up row-, column-, and cell-level evidence as the question unfolds. Encoder-side table structure and generic interleaved visual chain-of-thought still do not bind each reasoning step to that structure.
We propose \textbf{\name}, an \textbf{Inter}leaved structure-aware framework for CoT reasoning over \textbf{Tab}le images, interleaves chain-of-thought with tool calls that crop structure-aligned table regions. First, we build \dataset, includes reasoning trajectories in which each step is tied to both a structural location and a bounding box. \name is trained in two stages: supervised structure-aware alignment (SSA) on \dataset teaches the model to interleave reasoning with structure-aligned crops, and active localization optimization (ALO) further optimizes answer correctness, localization IoU, and output format, while penalizing missing or excessive tool calls. Experiments on nine table benchmarks show that \name improves the average accuracy of its backbone from 68.28\% to 73.17\% and achieves the best average performance among all compared methods. Code and data will be released soon.
\end{abstract}

\section{Introduction}

Tabular data is a fundamental format for organizing structural
information across real-world domains, and in many applications tables are available only as images, such as screenshots, and scanned reports\citep{huang2025hyperg,huang2025structfact}. Presenting tables as images also preserves layout and formatting cues, such as hierarchical headers, merged cells, and color shading, which are often lost or distorted when tables are serialized into text. Reasoning over a table image requires reading content organized by an explicit row-column structure: answering a question typically involves locating the relevant rows, following them to the right columns, and reading or comparing specific cells. Multimodal Large Language Models (MLLMs) can percept table images directly, making such reasoning possible without an external OCR or parsing pipeline.

\begin{figure}[t!]
    \centering
    \includegraphics[width=\textwidth]{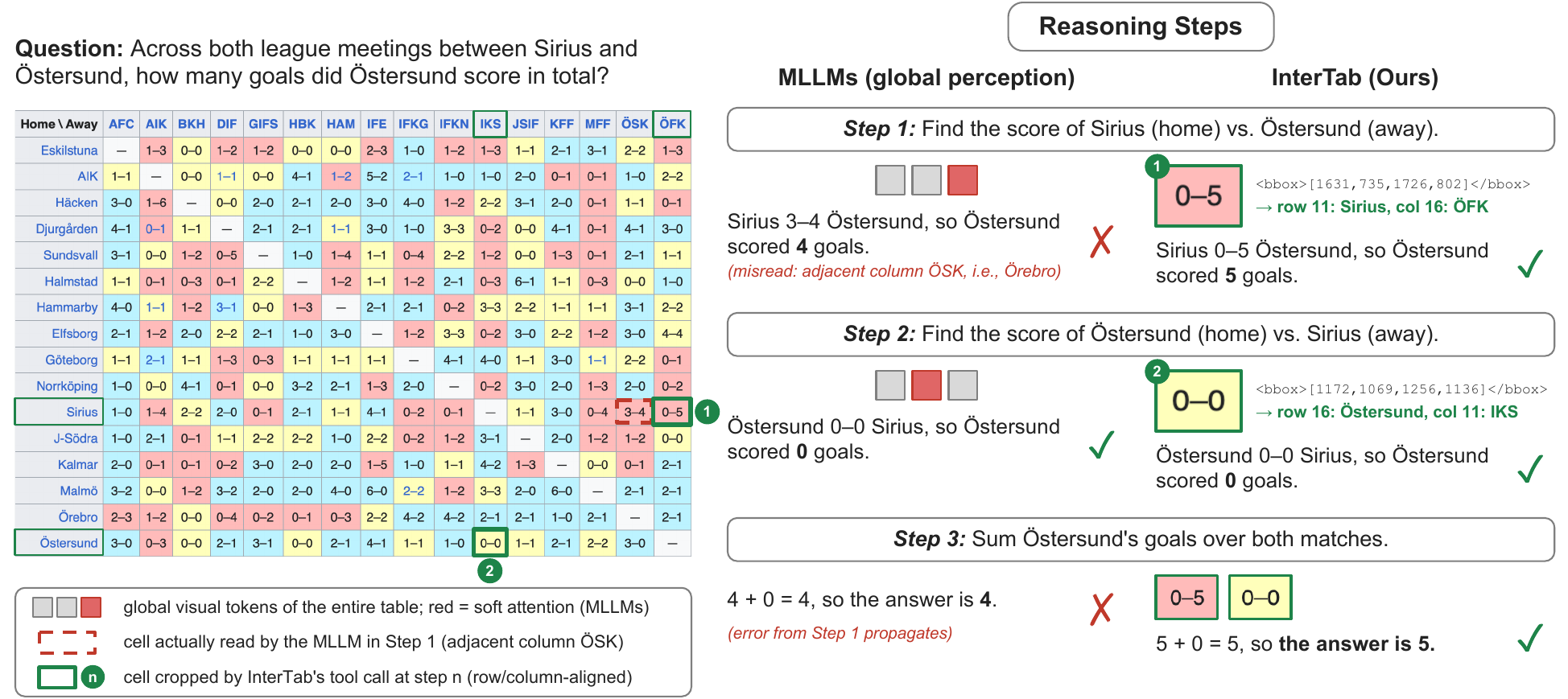}
    \caption{Comparison of two CoT-based reasoning methods on a table image.}
    \label{fig:1_toy}
\end{figure}

However, the visual perception of current MLLMs is fixed before reasoning begins: the whole image is encoded once in a single forward pass, \eg by CLIP-style visual encoders, and then consumed by the language decoder during generation, and every subsequent reasoning step can only attend to this same set of tokens\citep{ling2026debias,ling2026seeing,xun2026rtv}. This is poorly suited to multi-step table reasoning, where each step depends on a different small region of a complex or dense layout~\citep{Turbo,zheng2024picture}. Figure~\ref{fig:1_toy} shows a $16\times16$ match-result table. To count the goals Östersund scored against Sirius, the model must read two cells, \ie \texttt{ÖFK} and \texttt{ÖSK}, whose column headers are abbreviated and adjacent. A standard MLLM with Chain-of-Thought (CoT)~\citep{wei2022chain} decomposes the question correctly, but reads the neighboring column \texttt{ÖSK} instead of \texttt{ÖFK} in the first step, and the error propagates to the final answer. The failure lies not in language reasoning but in visual grounding: the model names the target row and column correctly, yet cannot precisely perceive the corresponding subregion in the image. We refer to this as \emph{passive} perception, since the visual evidence available to the model does not change with what the current reasoning step requires. Table reasoning instead calls for \emph{active} evidence acquisition, where the model decides at each step which rows, columns, or cells to inspect and obtains a precise view of them.

Existing methods for multimodal table reasoning mainly improve how the table image is encoded, either by distilling table understanding through supervised fine-tuning~\citep{Turbo,table-llava} or by injecting row and column structure into the visual encoder~\citep{CoCoTab,aida2025enhancing,zhang2025tablemoe}. Both still answer from a single global encoding. Another line of work, often called thinking with images, interleaves visual evidence with CoT on general images by selecting boxes~\citep{meng2025open,zheng2026deepeyes,he2025reasoning,su2026pixel} or querying patches~\citep{chen2026mintcot}. However, evidence in tables is inherently discrete: each value occupies a single cell defined by its row and column, so a crop shifted by just one cell can return an entirely different value. What is missing is a localization action that follows each reasoning step and is trained to land on the exact rows, columns, or cells that the step refers to.

To this end, we propose \textbf{\name}, which interleaves CoT with structure-aligned visual evidence. At each step, the model can call a \texttt{get\_sub\_table} tool to crop a table element, and the selected subregion is encoded as visual tokens that condition the subsequent reasoning. To enable such structure-aligned localization, we build \textbf{\dataset}, a set of 21.7K multi-turn trajectories in which each interleaved visual subregion is aligned, via table metadata, with the exact row, column, or cell referenced by the corresponding textual reasoning step. \name is trained in two stages: Supervised Structure-aware Alignment (SSA) fine-tunes the model on these trajectories, and Active Localization Optimization (ALO) applies Group Relative Policy Optimization (GRPO) with rewards for answer correctness, format, and localization IoU, together with a penalty on excessive tool calls.

We summarize our main contributions as follows.
\par\noindent\ding{182} We identify passive global perception as a key bottleneck in multimodal table reasoning, and formulate it as interleaved CoT with active, structure-aligned evidence acquisition.
\par\noindent\ding{183} We construct \dataset, to our knowledge the first interleaved multimodal table reasoning dataset, comprising 21.7K trajectories, in which the crop of each tool call is aligned with the rows, columns, or cells referenced by its reasoning step through metadata-guided correction.
\par\noindent\ding{184} We propose \textbf{\name}, an interleaved reasoning framework trained in two stages: SSA, which aligns each reasoning step with structure-aware visual evidence via supervised fine-tuning on \dataset, and ALO, which optimizes when and where to crop using the task and tool rewards.
\par\noindent\ding{185} We conduct extensive experiments on nine table datasets. \name improves the average accuracy of its backbone from 68.28 to 73.17 and achieves the best average performance among all compared methods in different table tasks.

Overall, \name shifts multimodal table reasoning from passive global encoding toward active, structure-aware evidence acquisition, providing a more reliable reasoning over visual tables.

\section{Related Works}
\par\noindent\textbf{Multimodal Table Reasoning.} Recent research extend analysis on textual tables to multimodal settings by leveraging document understanding systems or multimodal large language models (MLLMs) to jointly process visual and textual signals~\citep{nunes-etal-2025-benchmarking,yang2025does,10.5555/3692070.3692953}. Early work leverage OCR to firstly extract textual tables from the image and then processed for reasoning~\citep{8978013}. More recent approaches directly leverage end-to-end vision-language models (\eg represented by MLLMs) and their language reasoning capability~\citep{table-llava,zhang2025tablemoe,CoCoTab,lu2025ovis2}. For example, Turbo bridges the modality gap between structured tables and table images by leveraging privileged structured information and iterative reasoning optimization~\citep{Turbo}. Additionally, multimodal table and chart benchmarks~\citep{mathur-etal-2024-knowledge,10.1145/3774904.3792367,lompo2025visual} further advance multimodal reasoning capabilities in structured visual environments. Existing multimodal table reasoning still relies on a global view of the image, without binding each reasoning step to a row, column, or cell. Interleaved visual reasoning supplies that binding on natural images, but not under table structure. \name combines the two: the model reasons in text and, when needed, crops structure-aligned regions, learning to optimize where to crop rather than merely imitating demonstrations.
\par\noindent\textbf{Multimodal Interleaved Reasoning.} Multimodal interleaved reasoning equips models with the ability to generate alternating sequences of textual and visual tokens, enabling tighter coupling between visual perception and language-based reasoning \citep{chen2026mintcot,nie2026towards,gu2026thinkmorph,jin2026unveiling}. This paradigm improves cross-modal alignment by allowing intermediate reasoning steps to explicitly incorporate visual evidence. For instance, DeepEyes demonstrates that reinforcement learning–driven active perception can naturally elicit interleaved visual-text reasoning, leading to more grounded multimodal inference \citep{zheng2026deepeyes}. Such approaches further enhance chain-of-thought reasoning in multimodal settings, as evidenced by ICoT, which shows that interleaving visual and textual steps via a simple attention-based mechanism improves both performance and interpretability in MLLMs \citep{gao2025interleaved}.
\section{Preliminary}
\textbf{Table Image Reasoning with MLLMs.} Given a table image $I$ whose cells are arranged in rows and columns, a textual question $Q$, and an instruction prompt (Appendix ~\ref{sec:data_prompts}), an MLLM encodes $I$ into visual tokens $v_{1:N}$ and generates the output autoregressively:
\begin{equation}
p(x_{1:T}\mid I,Q)=\prod_{t=1}^{T} p(x_t \mid v_{1:N}, Q, x_{<t}).
\label{eq:pre_1}
\end{equation}
\textbf{CoT and Interleaved CoT.} Under the CoT setting, the output consists of reasoning steps $s_{1:k}$ followed by the final answer $y$, \eg first locating relevant rows and then narrowing down to target cells, yet every step still relies only on the global visual tokens $v_{1:N}$ encoded once from the whole image. Interleaved CoT additionally allows each step $s_j$ to end with a bounding box $b_j$, which represents the local visual evidence $\tilde{v}_j=\mathcal{E}_{vis}(\mathrm{crop}(I,b_j))$ ($\tilde{v}_j=\emptyset$ if no visual evidence is needed):
\begin{equation}
p(y, s_{1:k}\mid I,Q)=\prod_{j=1}^{k} p(s_j \mid v_{1:N}, Q, s_{<j}, \tilde{v}_{<j})\; p(y \mid v_{1:N}, Q, s_{1:k}, \tilde{v}_{1:k}).
\label{eq:pre_2}
\end{equation}
For table images, useful evidence is discrete and defined by row-column structure, so each $b_j$ must precisely cover the rows, columns, or cells referenced by $s_j$, which requires the structure-aware alignment introduced in Section~\ref{sec:method}.




\section{Methodology}
We first build \dataset, whose reasoning trajectories interleave textual steps with visual evidence (Section~\ref{sec:data_pipeline}). \name is then trained in two stages (Section~\ref{sec:method}): Supervised Structure-aware Alignment (SSA) , followed by Active Localization Optimization (ALO).

\begin{figure}[t!]
    \centering
    \includegraphics[width=\textwidth]{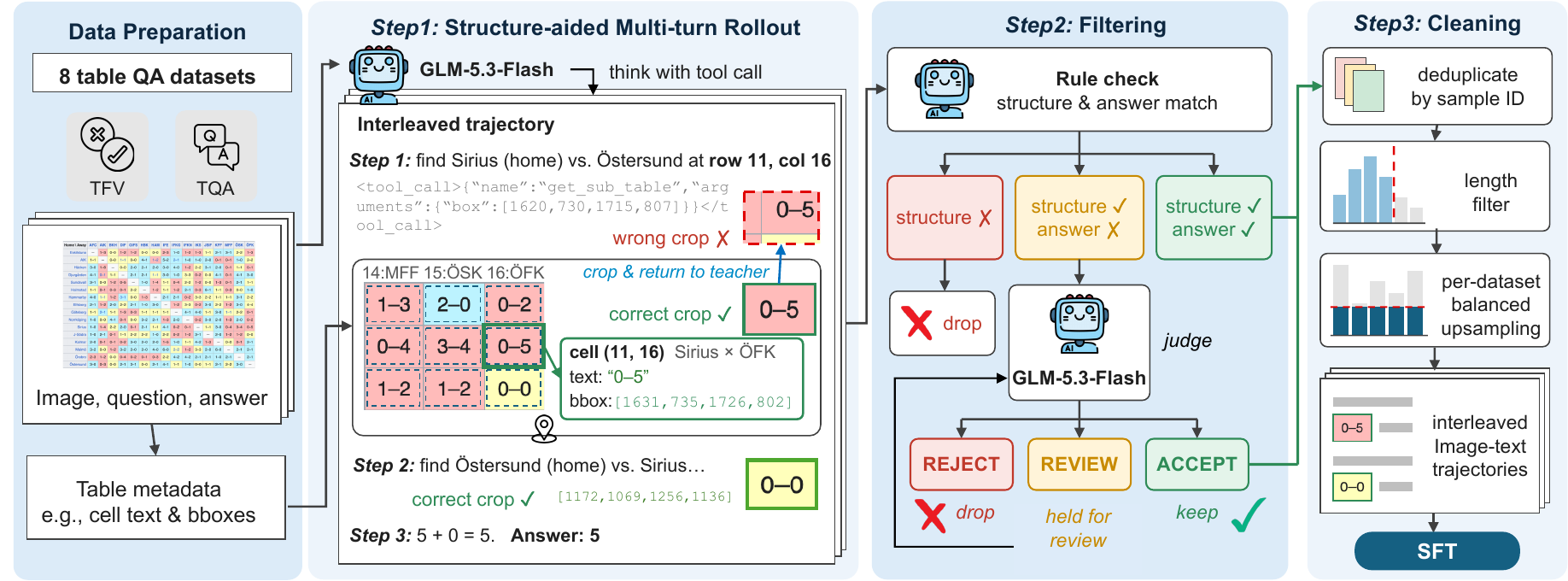}
    \caption{Construction pipeline of \dataset. \textbf{\textit{Step 1:}} GLM-5.3-Flash is instructed to generate multi-turn trajectories with \texttt{get\_sub\_table}, where wrong visual evidence are corrected by locating the referenced cell in the table metadata. \textbf{\textit{Step 2:}} LLM-as-judge for filtering trajectories based on visual-evidence structure and answer correctness. \textbf{\textit{Step 3:}} deduplicate trajectories by sample IDs, do length-filtered, and balanced across the eight source datasets for SFT.}
    \label{fig:data}
\end{figure}

\subsection{Structure-aided Reasoning Data Construction}\label{sec:data_pipeline}
We build \dataset from eight table datasets, \ie HiTab~\citep{cheng2022hitabhierarchicaltabledataset}, InfoTabs~\citep{gupta2020infotabsinferencetablessemistructured}, TabFact~\citep{Chen2020TabFact}, TabMWP~\citep{lu2023dynamicpromptlearningpolicy}, TAT-QA~\citep{zhu2021tatqaquestionansweringbenchmark}, WikiTableQA~\citep{pasupat-liang-2015-compositional}, FeTaQA~\citep{nan2021fetaqafreeformtablequestion}, and HybridQA~\citep{chen2021hybridqadatasetmultihopquestion}, from TABLET~\citep{alonso2026tabletlargescaledatasetrobust}, where each sample provides a table image, a question, its answer, and table metadata including the text and bounding box of every cell. As shown in Figure~\ref{fig:data}, the data is constructed in three steps.
\textbf{\textit{Step 1:} Structure-aided multi-turn rollout.} Given the raw table image and the question, we instruct GLM-5.3-Flash to reason step by step using the tool \texttt{get\_sub\_table}, which crops the region it needs as visual evidence at each step. On complex tables, however, the model tends to crop regions that are misaligned with cell boundaries. To address this, we identify the referenced cell from the textual reasoning step by its row and column, replace the proposed box with the exact bounding box from the table metadata, and return the corrected crop to the model to continue reasoning.
\textbf{\textit{Step 2:} Validation.} Then, each trajectory is validated by rule-based checks, which verify its structure (\eg well-formed tool calls and cropped visual regions) and compare the final answer with the ground truth. Trajectories that pass both checks are kept and those with invalid structure are discarded, while those with valid structure but mismatched answers are re-examined by a GLM-5.3-Flash judge, and only the accepted ones are kept. The prompts used are detailed in Appendix~\ref{sec:data_prompts}.
\textbf{\textit{Step 3:} Cleaning.} Finally, we deduplicate trajectories by the sample IDs and remove overly long ones, yielding 21,721 unique trajectories. For SFT, the eight source datasets are balanced by upsampling.

\subsection{\name}\label{sec:method}
Building on \dataset, we propose \name, a two-stage framework consisting of Supervised Structure-aware Alignment (SSA) and Active Localization Optimization (ALO). Specifically, SSA performs cold start to incrementally align visual and structural information, while ALO further trains the model via tool-augmented to improve reasoning accuracy and visual localization quality.
\subsubsection{Supervised Structure-aware Alignment (SSA)}
SSA fine-tunes the MLLM on \dataset, $\mathcal{D}=\{(I_i, Q_i, y_i, s^{(i)}_{1:k})\}$. In step $s^{(i)}$ that needs visual evidence, the reasoning texts is interleaved with a structure-aligned bounding box $b_j$ is aligned with the rows, columns, or cells referred in that step. The bounding visual subregion is encoded as visual tokens $\tilde{v}_j$ and appended to the reasoning, as in Eq.~\eqref{eq:pre_2}. We minimize the negative log-likelihood
\begin{equation}
\mathcal{L}_{\text{SSA}}(\theta)
= - \sum_{(I, Q, y, s_{1:k}) \in \mathcal{D}}
\log p_{\theta}(y, s_{1:k} \mid I, Q),
\end{equation}

to enable the MLLM to locate the correct visual subregions within the entire table image $I$.

\subsubsection{Active Localization Optimization (ALO)}
SSA teaches the model where visual evidence locate, but does not optimize when and where to crop under its own reasoning. At the stage of ALO, we set evidence acquisition as part of the learning policy: each tool call and its box are actions sampled by the model and rewarded by their outcomes. Specifically, ALO can be either applied upon the stage of SSA or directly to the base model, using reward formualted in \ref{eq:reward_function}. For each $(I, Q)$, we sample $G$ responses $\{o_i\}_{i=1}^{G}$ from $\pi_{\theta_{\text{old}}}$ and maximize
\begin{equation}
\begin{aligned}
\mathcal{J}_{\text{ALO}}(\theta) = \mathbb{E}\bigg[ \frac{1}{G} \sum_{i=1}^{G} \Big( & \min \big( \rho_i(\theta) A_i,\ \text{clip}(\rho_i(\theta), 1-\epsilon, 1+\epsilon) A_i \big) \\
& - \beta D_{\text{KL}}(\pi_{\theta} \,\|\, \pi_{\text{ref}}) \Big) \bigg],
\end{aligned}
\end{equation}
where $\rho_i(\theta) = \pi_{\theta}(o_i \mid I, Q) / \pi_{\theta_{\text{old}}}(o_i \mid I, Q)$ and $A_i = \big(R_i - \mathrm{mean}(\{R_j\}_{j=1}^{G})\big) / \mathrm{std}(\{R_j\}_{j=1}^{G})$.

\textbf{Reward.} The reward of a response combines task and tool terms:
\begin{equation}
R = \lambda_{\text{acc}} R_{\text{acc}} + \lambda_{\text{fmt}} R_{\text{fmt}} + \lambda_{\text{tool}} R_{\text{tool}} - \lambda_{\text{miss}}\, \mathbb{I}_{\text{miss}} - P_{\text{tool}}.
\label{eq:reward_function}
\end{equation}
$R_{\text{acc}}$ is $1$ on exact match with $y$; otherwise, an LLM judge returns $1$ or $0$. $R_{\text{fmt}}$ is $1$ when both the thinking span and a complete answer tag are present, $0.5$ when only the answer tag is present, and $0$ otherwise. $R_{\text{tool}}$ averages, over the ground-truth evidence boxes $\mathcal{B}^{*}$ from the table metadata, the best IoU achieved by the predicted boxes $\mathcal{B}$:
\begin{equation}
R_{\text{tool}} = \frac{1}{|\mathcal{B}^{*}|} \sum_{b^{*} \in \mathcal{B}^{*}} \max_{b \in \mathcal{B}} \mathrm{IoU}(b, b^{*}),
\end{equation}
and $R_{\text{tool}} = 0$ if there are no ground-truth boxes or no predicted boxes, \ie $\mathcal{B}^{*} = \emptyset$ or $\mathcal{B} = \emptyset$. $\mathbb{I}_{\text{miss}} = 1$ if $\mathcal{B}^{*} \neq \emptyset$ but the response makes no tool call. The tool-use penalty $P_{\text{tool}} = \min\big(\lambda_{\text{pen}} \max(N - N_0, 0),\, P_{\max}\big)$ penalizes calls beyond the first $N_0$, where $N$ is the number of tool calls. 

\begin{figure}[t!]
    \centering
    \includegraphics[width=\textwidth]{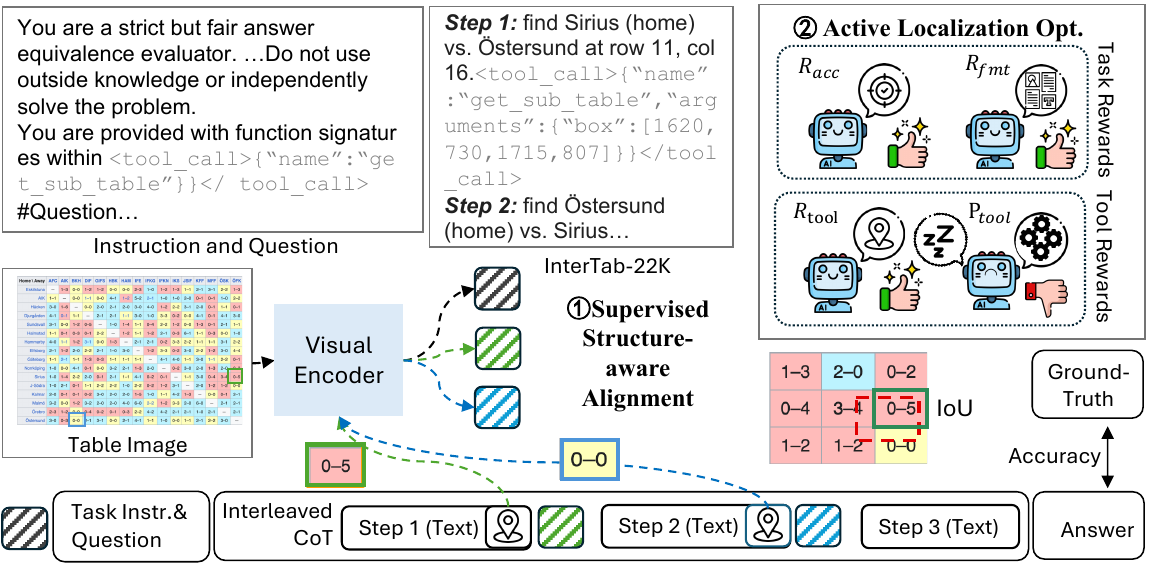}
    \caption{Overview of the \textbf{\name} framework. First, we perform supervised alignment of each reasoning step with interleaved, structure-aware visual tokens using \dataset. Second, we optimize the model using ALO with tool and task rewards, enabling an active policy for interleaved reasoning and accurate visual localization.} 
    \label{fig:method}
\end{figure}
\section{Experiment}
We organize our experiments around four research questions: \noindent\ding{182}How does \name compare with existing models across in-domain table reasoning benchmarks? \noindent\ding{183}Does interleaving textual reasoning with structure-aligned visual evidence improve accuracy over text-only reasoning? \noindent\ding{184}How do the training stages and auxiliary reward terms affect performance, and how sensitive is \name to the tool-call allowance? \noindent\ding{185}Does \name retain its performance advantage on a held-out benchmark excluded from the training data?

\subsection{Experimental Settings} \label{sec:settings}
We use Qwen3-VL-8B-Instruct as the base model and train on 8 NVIDIA A100 GPUs. For SSA, we set the batch size to 64, the learning rate of $1 \times 10^{-5}$, and train for 1 epoch. For ALO, we set the batch size to 16, the learning rate to $1 \times 10^{-6}$, the rollout number $n$ to 8, and train for 2 epoch. For evaluation, we use nine representative benchmark: HiTab, InfoTabs, TabFact, TabMWP, TATQA, WikiTQ, FeTaQA, HybridQA, and VisualTQA~\citep{lompo2025visual} (out-of-domain), with the in-domain data taken from TABLET~\citep{alonso2026tabletlargescaledatasetrobust}. To accurately evaluate the models' responses, we conduct an LLM-as-a-judge evaluation (see Appendix for details).

\subsection{Main Results}\label{sec:main_res}
\begin{table*}[t]
    \centering
    \caption{
        Performance comparison on nine table understanding benchmarks. Accuracy (\%) is reported for all benchmarks. Best results are in bold and second-best results are underlined.
    }
    \label{tab:main_result}
    \resizebox{\textwidth}{!}{
    \begin{tabular}{@{}lcccccccccc@{}}
        \toprule
        Model
        & HiTab
        & InfoTabs
        & TabFact
        & TabMWP
        & TATQA
        & WikiTQ
        & FeTaQA
        & HybridQA
        & VTQA
        & Average \\
        \midrule

        \multicolumn{11}{l}{\textit{General-purpose MLLMs}} \\

        GLM-4.1V-9B~\citep{vteam2026glm45vglm41vthinkingversatilemultimodal}
        & 71.43 & \underline{72.93} & 82.03 & 86.58
        & 38.37 & \underline{72.03} & 38.81
        & 54.01 & 51.75 & 63.10 \\

        InternVL3.5-8B~\citep{wang2025internvl35advancingopensourcemultimodal}
        & \underline{77.12} & 71.67 & \underline{87.10} & \underline{95.53}
        & \underline{51.26} & 69.07
        & \underline{42.17} & 70.94
        & 49.67 & \underline{68.28} \\

        Kimi-VL-A3B-Instruct~\citep{kimiteam2025kimivltechnicalreport}
        & 62.46 & 66.76 & 70.43 & 54.72
        & 16.04 & 56.99 & 38.92
        & 48.73 & 38.11 & 50.35 \\

        LLaVA-1.6~\citep{liu2024llavanext}
        & 16.84 & 37.81 & 49.78 & 39.34
        & 9.03 & 16.64 & 13.33
        & 9.93 & 21.24 & 23.77 \\

        Qwen2.5-VL-7B-Instruct~\citep{bai2025qwen25vltechnicalreport}
        & 73.86 & 71.65 & 76.00 & 89.84
        & 40.78 & 60.14 & 39.20
        & 39.87 & 46.64 & 59.78 \\

        Qwen3-VL-8B-Instruct~\citep{bai2025qwen3vltechnicalreport}
        & 74.36 & 72.19 & 82.25
        & 95.05 & 47.65
        & 68.67 & 41.99 & 69.22
        & \underline{53.27} & 67.18 \\

        \midrule
        \multicolumn{11}{l}{\textit{Table-specialized Models}} \\

        TableLLaVA-7B~\citep{table-llava}
        & 9.11 & 57.54 & 60.14 & 61.42
        & 15.69 & 17.52 & 20.42
        & 19.64 & 25.19 & 31.85 \\

        TableVision-R1~\citep{tablevisionr1}
        & 63.68 & 62.98 & 72.17 & 83.72
        & 29.41 & 50.72 & 34.04
        & 37.05 & 33.88 & 51.96 \\

        MiniCPM-V-2.6~\citep{yao2024minicpmvgpt4vlevelmllm}
        & 48.04 & 56.56 & 73.89 & 87.98
        & 24.35 & 57.92 & 29.27
        & 22.50 & 30.39 & 47.88 \\

        HIPPO~\citep{wang2025hippoenhancingtableunderstanding}
        & 46.37 & 64.39 & 74.68 & 86.85
        & 24.44 & 56.39 & 28.63
        & 25.39 & 31.05 & 48.69 \\

        Ovis2~\citep{lu2024ovisstructuralembeddingalignment}
        & 66.09 & 71.04
        & 82.28 & 92.62
        & 42.96 & 57.49 & 41.88
        & \underline{77.32} & 45.76
        & 64.16 \\

        \midrule
        \multicolumn{11}{l}{\textit{Thinking with Images}} \\

        DeepEyes~\citep{zheng2026deepeyes}
        & 71.80 & 70.17 & 77.07 & 92.97
        & 34.43 & 60.88 & 40.78
        & 46.75 & 47.21 & 60.23 \\

        MINT-COT~\citep{chen2026mintcot}
        & 51.01 & 59.54 & 67.94 & 78.75
        & 36.23 & 38.62 & 31.63
        & 24.05 & 25.82 & 45.95 \\

        \midrule
        \rowcolor{gray!15}
        \textbf{InterTab (Ours)}
        & \textbf{77.45} & \textbf{73.59} & \textbf{87.95} & \textbf{96.35}
        & \textbf{57.25} & \textbf{75.03} & \textbf{50.10}
        & \textbf{81.00} & \textbf{59.80} & \textbf{73.17} \\
        \bottomrule
    \end{tabular}
    }
\end{table*}

To address RQ1, we compare \name with existing models on the eight in-domain benchmarks. Table~\ref{tab:main_result} shows that \name achieves the best accuracy on all eight in-domain benchmarks, consistently outperforming the existing baselines. Compared with the Qwen3-VL-8B backbone, \name improves HiTab \citep{cheng2022hitabhierarchicaltabledataset}, InfoTabs \citep{gupta2020infotabsinferencetablessemistructured}, TabFact \citep{Chen2020TabFact}, TabMWP \citep{lu2023dynamicpromptlearningpolicy}, TATQA \citep{zhu2021tatqaquestionansweringbenchmark}, WikiTQ \citep{pasupat-liang-2015-compositional}, FeTaQA \citep{nan2021fetaqafreeformtablequestion}, and HybridQA~\citep{chen2021hybridqadatasetmultihopquestion} by 3.09, 1.40, 5.70, 1.30, 9.60, 6.36, 8.11, and 11.78 percentage points, respectively. The largest gains are observed on HybridQA, TATQA, and FeTaQA, where \name improves over the backbone by more than 8 points.
The magnitude of improvement varies across benchmarks. On TabMWP and InfoTabs, where the backbone already performs strongly, \name still yields gains of 1.30 and 1.40 points, respectively. The improvements increase to 3.09 and 5.70 points on HiTab and TabFact, and reach 9.60, 6.36, 8.11, and 11.78 points on TATQA, WikiTQ, FeTaQA, and HybridQA, respectively. Overall, \name achieves a mean improvement of 5.92 points across the eight in-domain benchmarks, while its nine-benchmark average reaches 73.17\%, compared with 67.18\% for the Qwen3-VL-8B backbone.
We further compare \name with both table-specialized and general-purpose vision-language models. Among the table-specialized baselines, Ovis2 achieves an average accuracy of 64.16v across the nine benchmarks, while InternVL3.5-8B achieves 68.28\% and represents the strongest competing baseline. Thinking-with-images methods, including DeepEyes and MINT-CoT, obtain averages of 60.23\% and 45.95\%, respectively, both below the Qwen3-VL-8B backbone. These comparisons show that \name consistently outperforms existing approaches across all eight training benchmarks, achieving the strongest overall performance among the evaluated methods rather than relying on gains from a single benchmark or model family.

\subsection{Further Study}\label{sec:ablation}
To address RQ2 and RQ3, we examine the design of \name in four steps: the reasoning paradigm, the two-stage training strategy, the reward components, and the tool-call allowance. The first three comparisons assess the contribution of each design choice; the last tests sensitivity to the call-penalty threshold. Unless otherwise stated, we report average accuracy across all nine benchmarks, including held-out VTQA. All accuracy differences below are in percentage points.

\subsubsection{Interleaved Reasoning}\label{sec:reasoning_ablation}
We first test whether interleaving reasoning with localized visual evidence improves accuracy. To this end, we establish a matched text-only baseline using the same Qwen3-VL-8B-Instruct backbone and training recipe. We convert our interleaved training data into a text-only version by retaining the same questions and answer targets while removing localization calls and the corresponding cropped table regions. As shown in Table~\ref{tab:reasoning_strategy}, interleaved reasoning consistently outperforms text-only reasoning across all nine benchmarks, improving the average from 68.89\% to 73.17\% (+4.28 points). The gains are particularly large on HybridQA (+8.39 points), HiTab (+6.00 points), and TATQA (+6.00 points), and remain substantial on held-out VTQA (+5.41 points). These results demonstrate the effectiveness of interleaved reasoning over text-only reasoning.

\begin{table*}[t]
\centering
\caption{Interleaved reasoning versus a matched text-only baseline. Accuracy (\%) is reported.}
\label{tab:reasoning_strategy}
\setlength{\tabcolsep}{3pt}
\renewcommand{\arraystretch}{1.12}
\resizebox{\textwidth}{!}{%
\begin{tabular}{lcccccccccc}
\toprule
Reasoning & HiTab & InfoTabs & TabFact & TabMWP & TATQA & WikiTQ & FeTaQA & HybridQA & VTQA & Average \\
\midrule
Text-only CoT
& 71.45 & 70.44 & 87.01 & 94.91 & 51.25 & 72.78 & 45.19 & 72.60 & 54.40 & 68.89 \\
Interleaved (Ours)
& \textbf{77.45} & \textbf{73.59} & \textbf{87.95} & \textbf{96.35} & \textbf{57.25} & \textbf{75.03} & \textbf{50.10} & \textbf{81.00} & \textbf{59.80} & \textbf{73.17} \\
\rowcolor{gray!15}
$\Delta$
& +6.00 & +3.15 & +0.94 & +1.44 & +6.00 & +2.25 & +4.91 & +8.39 & +5.41 & +4.28 \\
\bottomrule
\end{tabular}%
}
\end{table*}

\subsubsection{Two-Stage Training}\label{sec:training_ablation}
\begin{figure}[t!]
    \centering
    \includegraphics[width=\textwidth]{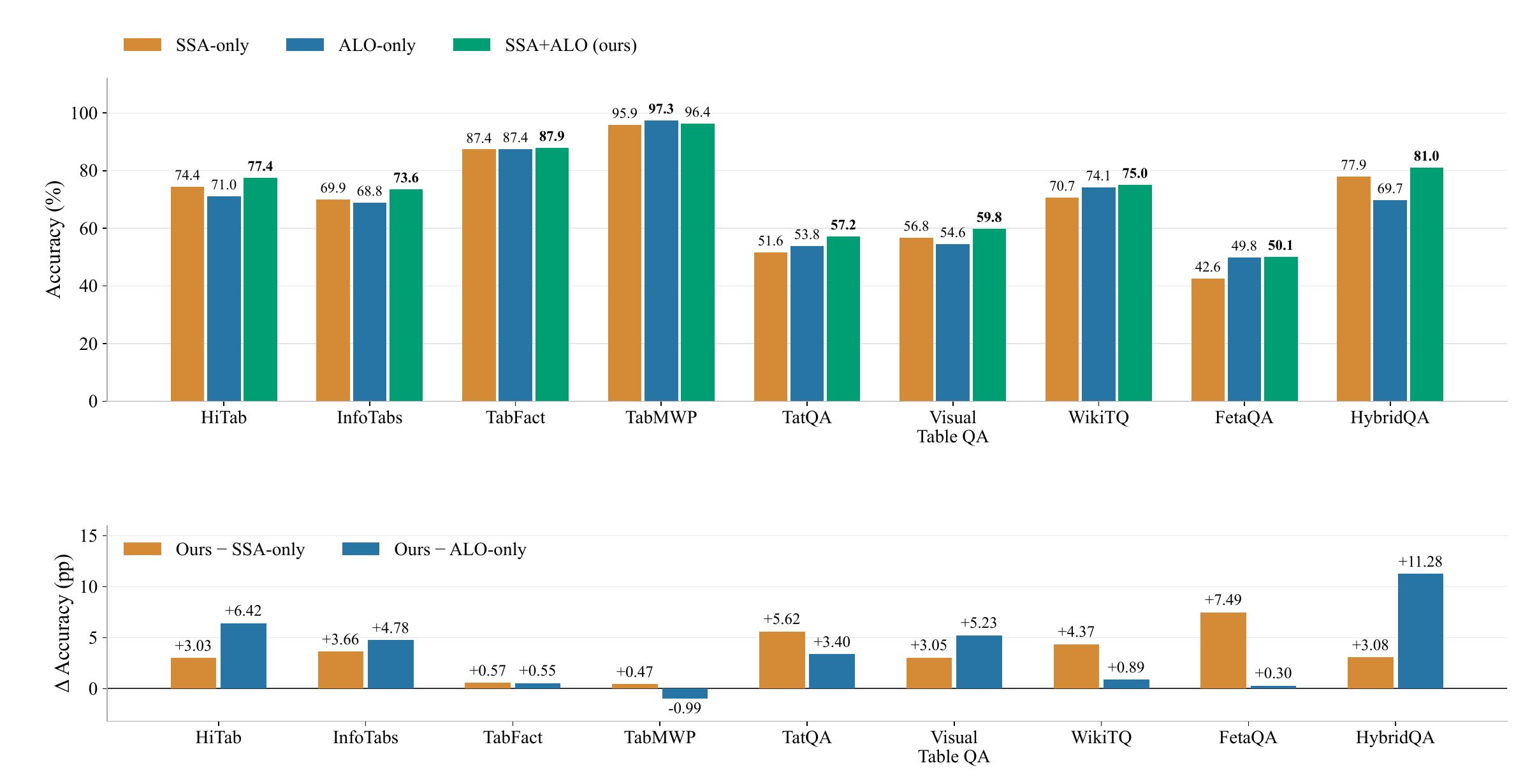}
    \caption{Training-strategy comparison. Top: accuracy of SSA-only, ALO-only, and SSA+ALO. Bottom: accuracy gain of SSA+ALO over each baseline, in percentage points.}
    \label{fig:training_strategy}
\end{figure}

Having established the benefit of interleaved reasoning, we examine how the two training stages contribute. Specifically, we train three variants from the same backbone on the same interleaved data: SSA-only, ALO-only, and SSA followed by ALO (Ours). As shown in Figure~\ref{fig:training_strategy}, SSA+ALO reaches an average of 73.17\%, above SSA-only (69.69\%) and ALO-only (69.63\%), by 3.48 and 3.54 points, and is best on eight of the nine benchmarks. Relative to SSA-only, the largest gains are on FeTaQA (+7.49 points), TATQA (+5.62 points), and WikiTQ (+4.37 points). Relative to ALO-only, the largest gains are on HybridQA (+11.28 points), HiTab (+6.42 points), and VTQA (+5.23 points). The exception is TabMWP, where ALO-only achieves 97.34\% compared with 96.35\% for SSA+ALO. We attribute this to the simpler table structures in TabMWP, which make direct ALO sufficient for learning high-reward solutions. These results show that the two stages are complementary: SSA establishes the interleaved localization behavior, while ALO further improves it, and neither stage alone matches the joint recipe.

\subsubsection{Reward Components}\label{sec:reward_ablation}
To further analyze the contribution of each auxiliary reward term, we conduct an ablation study within the full two-stage recipe. Specifically, we remove each auxiliary term from Eq.~\eqref{eq:reward_function} while keeping answer correctness and format rewards unchanged. As shown in Table~\ref{tab:reward_ablation}, removing the localization reward reduces the average performance by 2.87 points to 70.30\%, while removing the call penalty reduces it by 1.69 points to 71.47\%. Both variants underperform the full reward on eight of nine benchmarks. The localization reward is particularly beneficial for precise evidence selection, with drops of 9.15 points on FeTaQA, 5.26 on TATQA, and 4.87 on InfoTabs. The call penalty also yields notable gains on FeTaQA and InfoTabs, indicating that it helps suppress unnecessary tool calls.
TabMWP is the main exception: removing the localization reward improves performance from 96.35\% to 97.46\%. We attribute this to its relatively simple and compact tables, where precise localization is often unnecessary for obtaining the answer. The same trend appears in the training-strategy comparison, where ALO-only slightly outperforms SSA+ALO on TabMWP (97.34\% vs. 96.35\%). Overall, the results support retaining both auxiliary terms, as their gains on localization- and tool-use-sensitive benchmarks outweigh the small degradation on TabMWP.

\begin{table*}[t]
\centering
\caption{Reward-component ablation. Accuracy (\%) is reported. $\Delta$ is the change relative to Ours.}
\label{tab:reward_ablation}
\setlength{\tabcolsep}{3pt}
\renewcommand{\arraystretch}{1.12}
\resizebox{\textwidth}{!}{%
\begin{tabular}{lcccccccccc}
\toprule
Configuration & HiTab & InfoTabs & TabFact & TabMWP & TATQA & WikiTQ & FeTaQA & HybridQA & VTQA & Average \\
\midrule
Ours & \textbf{77.45} & \textbf{73.59} & \textbf{87.95} & 96.35 & \textbf{57.25} & \textbf{75.03} & \textbf{50.10} & \textbf{81.00} & 59.80 & \textbf{73.17} \\
\midrule
w/o call penalty & 76.39 & 70.70 & 86.59 & 94.80 & 54.87 & 74.13 & 43.03 & 80.01 & \textbf{62.75} & 71.47 \\
\rowcolor{gray!15}
$\Delta$ & $-1.06$ & $-2.89$ & $-1.36$ & $-1.55$ & $-2.38$ & $-0.90$ & $-7.07$ & $-0.99$ & $+2.95$ & $-1.69$ \\
\midrule
w/o localization & 76.14 & 68.72 & 86.02 & \textbf{97.46} & 51.99 & 73.93 & 40.95 & 80.59 & 56.86 & 70.30 \\
\rowcolor{gray!15}
$\Delta$ & $-1.31$ & $-4.87$ & $-1.93$ & $+1.11$ & $-5.26$ & $-1.10$ & $-9.15$ & $-0.41$ & $-2.94$ & $-2.87$ \\
\bottomrule
\end{tabular}%
}
\end{table*}

\subsubsection{Tool-Call Allowance}\label{sec:allowance}
To further analyze the effect of the unpenalized allowance $N_0$, we conduct a sensitivity study by fixing the SSA+ALO recipe and varying $N_0$ from 3 to 9. As shown in panel (a, b) of Figure~\ref{fig:tool_call_sensitivity}, $N_0=7$ achieves the highest nine-benchmark average of 73.17, compared with 71.30\%, 71.58\%, and 72.99\% for $N_0=3,5,9$, respectively. Although different benchmarks favor different allowances, $N_0=7$ provides the best overall performance. We therefore fix $N_0=7$ in the main experiments to achieve more comprehensive reasoning performance across benchmarks.

\subsection{Out-of-Domain Evaluation}\label{sec:ood}
\begin{figure}[t!]
    \centering
    \includegraphics[width=\textwidth]{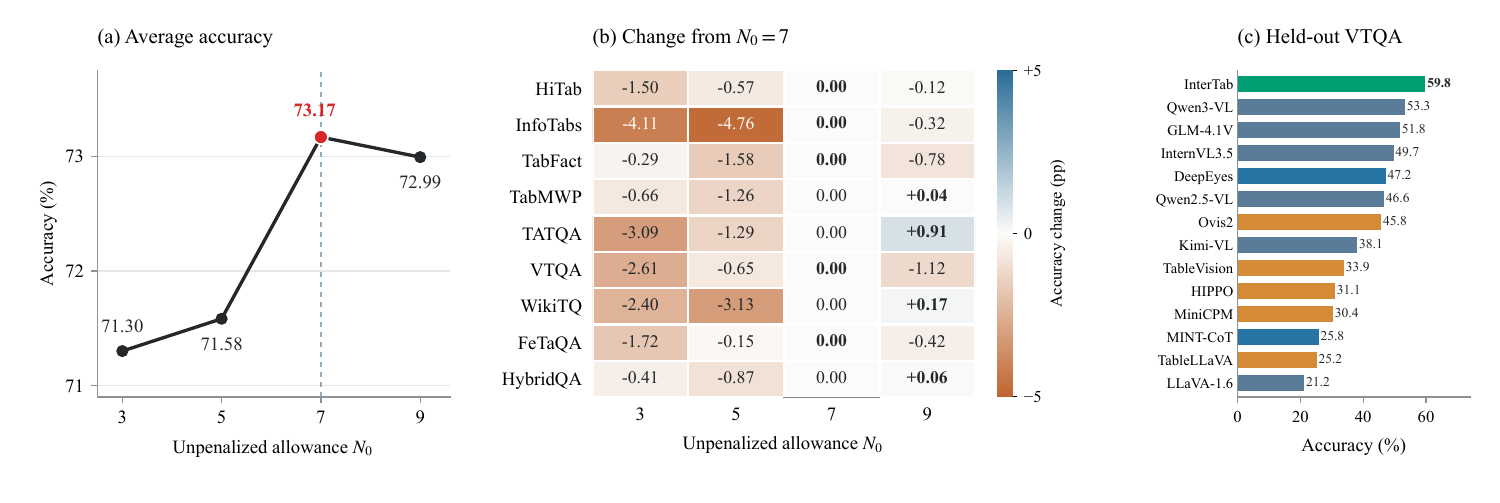}
    \caption{(a) Average accuracy across unpenalized allowances. (b) Per-benchmark change from $N_0=7$. (c) Accuracy on held-out Visual Table QA.}
    \label{fig:tool_call_sensitivity}
    \label{fig:vtqa_ood}
\end{figure}
To address RQ4, we evaluate every model in Table~\ref{tab:main_result} on Visual Table QA, which does not appear in \dataset. Panel (c) of Figure~\ref{fig:vtqa_ood} ranks these results. \name scores 59.80\%, ahead of the Qwen3-VL-8B backbone at 53.27\% (+6.53) and GLM-4.1V-9B at 51.75\%. Table-specialized models that are competitive on the in-domain benchmarks fall further behind: Ovis2 scores 45.76\%, and TableLLaVA scores 25.19\%. Thinking-with-images methods show the same gap. DeepEyes scores 47.21\% and MINT-CoT scores 25.82\%, both below the backbone. The ordering indicates that the gain is not a generic vision-language improvement. Cropping structure-aligned regions carries over to this unseen set, whereas encoding the whole table once, or cropping image regions without table structure, does not. Although removing the call penalty raises VTQA to 62.75\%, the full setting provides the best overall trade-off across benchmarks and is therefore retained as the default configuration.

\subsection{Case Study}\label{sec:case}
\begin{figure}[t!]
    \centering
    \includegraphics[width=\textwidth]{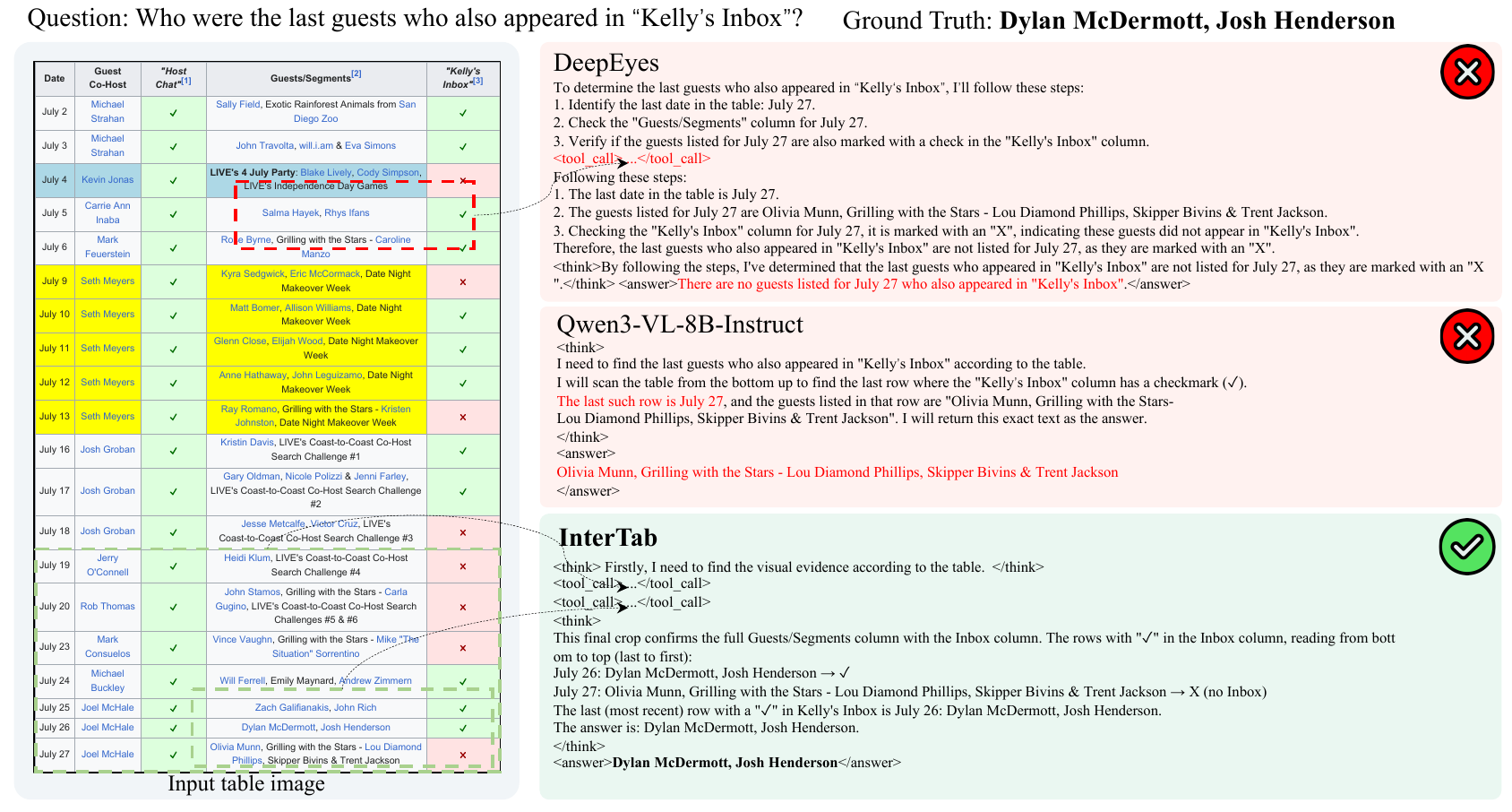}
    \caption{Case study of \name. Our method accurately localizes the most relevant region.}
    \label{fig:case_study}
\end{figure}
To complement the quantitative comparison in RQ2, Figure~\ref{fig:case_study} illustrates how structure-aligned visual evidence supports row selection. The task requires finding the latest row with a checked `Kelly's Inbox'' entry and retrieving the corresponding guests. DeepEyes and Qwen3-VL-8B-Instruct both select the wrong row, while InterTab accurately localizes the relevant rows and columns, identifies the unchecked entry on July 27, and selects the correct entry on July 26, returning `Dylan McDermott, Josh Henderson.'' This example demonstrates the effectiveness of InterTab's interleaved grounding paradigm for table reasoning.

\section{Conclusion}
In this paper, we present \textbf{\name}, which interleaves chain-of-thought reasoning with structure-aligned cropping for table reasoning. \textbf{\dataset} provides the training traces, while SSA establishes the format and ALO optimizes accurate localization under a unified reward. Across nine benchmarks, \name achieves strong overall performance, with ablations showing that SSA and ALO are complementary and that interleaved grounding becomes most effective after optimization.

\section*{AI Use Statement}

In this work, we used generative AI tools to assist with writing, including grammar checking and language polishing, and for general-purpose coding support. Generative AI was also used as part of the proposed research methodology: as described in Section~\ref{sec:data_pipeline}, GLM-5.3-Flash was used to construct the reasoning traces for \dataset. We did not use generative AI for research ideation, identifying or retrieving related work, or reviewing-related tasks. All AI-assisted work was reviewed and verified by the authors: AI-assisted text was checked and edited, AI-assisted code was tested, and the generated data were filtered and manually spot-checked. We take full responsibility for the final content of this work, including all text, claims, code, and data.

\section*{Ethics Statement}

This work uses publicly available table reasoning datasets. Training traces in \dataset come from HiTab, InfoTabs, TabFact, TabMWP, TATQA, WikiTableQuestions, FeTaQA, and HybridQA. Visual Table QA is used only for out-of-domain evaluation. The work does not involve human subjects, private data, or personally identifiable information. Our models may
still produce incorrect reasoning chains or inaccurately grounded references,
especially under ambiguous layouts or low-quality scans, so \name should be used
as an assistive reasoning tool rather than an autonomous decision maker in
high-stakes domains. A broader discussion of societal impacts, risks, and
responsible deployment is provided in Appendix~\ref{sec:impacts}.

\section*{Reproducibility Statement}

We have made the following efforts to ensure reproducibility. The construction of
\dataset, including the three-stage dual-aligned generation pipeline and the
filtering criteria, is described in Section~\ref{sec:data_pipeline}, with the complete set
of prompts in Appendix~\ref{sec:data_prompts} and dataset statistics in
Appendix~\ref{sec:data_stats}. The training objectives and the full definition of
every reward and penalty term are given in Section~\ref{sec:method} and
Appendix~\ref{appendix:acc_reward}--\ref{appendix:format_reward}. All backbone
models, hardware, batch sizes, learning rates, and sampling settings used in both
training stages are reported in Section~\ref{sec:settings}. Anonymized code and
data will be released.

\bibliography{ref}
\bibliographystyle{preprint}

\appendix


\section{Dataset Details}
In this section, we provide detailed statistics of our constructed dual-aligned dataset \dataset in Section \ref{sec:data_stats}, and present the prompts used in the generation pipeline in Section \ref{sec:data_prompts}.

\subsection{Dataset Statistics}\label{sec:data_stats}
Table~\ref{tab:data_stats} summarizes the key statistics of the \dataset training set used in our experiments.
The statistics are computed over 21,721 source records after length filtering and before upsampling. Balanced upsampling yields 53,048 examples per SSA epoch. Table~\ref{tab:dataset_composition} reports the source-wise composition; the ALO subset contains 3,200 training and 80 validation prompts before prompt-length filtering. The SSA-initialized ALO run retains 3,196 training prompts after filtering.

\begin{table}[htbp]
\centering
\caption{Statistics of the SSA source dataset before upsampling. Text tokens are counted using the Qwen3-VL tokenizer without chat-template framing or image-patch tokens; literal \texttt{<image>} placeholders are retained. Observations contain only image placeholders.}
\label{tab:data_stats}
\begin{tabular}{l|rrr}
\toprule
\textbf{Statistic} & \textbf{Mean} & \textbf{Min} & \textbf{Max} \\
\midrule
\multicolumn{4}{l}{\textit{General}} \\
Total source samples & \multicolumn{3}{c}{21,721} \\
Samples containing tool calls & \multicolumn{3}{c}{21,721 (100\%)} \\
Unique tools & \multicolumn{3}{c}{1 (\texttt{get\_sub\_table})} \\
Effective samples per SSA epoch & \multicolumn{3}{c}{53,048 (with upsampling)} \\
\midrule
\multicolumn{4}{l}{\textit{Conversation \& Reasoning}} \\
Assistant turns per sample$^{a}$ & 5.44 & 4 & 9 \\
Explicit reasoning blocks per sample$^{b}$ & 1.00 & 0 & 1 \\
\midrule
\multicolumn{4}{l}{\textit{Tool Calls}} \\
Tool calls per sample & 4.44 & 3 & 8 \\
Max calls in a single turn & 1.00 & 1 & 1 \\
Arguments per tool call & 1.00 & 1 & 1 \\
\midrule
\multicolumn{4}{l}{\textit{Text Token Length}} \\
Total tokens per sample & 2,297.24 & 314 & 31,872 \\
Input tokens per sample$^{c}$ & 1,933.17 & 146 & 31,652 \\
Function-call tokens per sample & 183.06 & 100 & 374 \\
Observation tokens per sample & 13.31 & 9 & 24 \\
Output tokens per sample & 167.70 & 21 & 4,670 \\
\bottomrule
\end{tabular}
\par\smallskip
\begin{minipage}{\linewidth}\footnotesize
$^{a}$Function-call messages plus final assistant messages; observations are excluded.
$^{b}$Non-empty \texttt{<think>} blocks, not semantic reasoning steps. One sample has no such block; the mean rounds to 1.00.
$^{c}$Human messages plus stored system and tool-schema text. Each message/field is tokenized separately with special-token insertion disabled. Statistics are unweighted across source records, not the upsampled mixture. Token maxima exclude visual patches and thus are not the full multimodal context lengths.
\end{minipage}
\end{table}

\begin{table}[t]
\centering
\caption{Dataset composition before and after SSA upsampling. ALO samples are drawn without replacement from the SSA source pool. ALO counts precede prompt-length filtering.}
\label{tab:dataset_composition}
\begin{tabular}{lrrrr}
\toprule
Dataset & SSA source & SSA/epoch & ALO train & ALO val. \\
\midrule
FeTaQA & 1,311 & 6,631 & 400 & 10 \\
HiTab & 1,216 & 6,631 & 400 & 10 \\
HybridQA & 6,513 & 6,631 & 400 & 10 \\
InfoTabs & 410 & 6,631 & 400 & 10 \\
TabFact & 6,631 & 6,631 & 400 & 10 \\
TabMWP & 4,380 & 6,631 & 400 & 10 \\
TAT-QA & 487 & 6,631 & 400 & 10 \\
WikiTQ & 773 & 6,631 & 400 & 10 \\
\midrule
Total & 21,721 & 53,048 & 3,200 & 80 \\
\bottomrule
\end{tabular}
\end{table}

\subsection{Training Hyperparameters}\label{sec:appendix_hyperparameters}
Table~\ref{tab:training_hyperparameters} lists the SSA configuration and the recorded settings of the two main ALO runs. The SSA schedule specifies three epochs, but the checkpoint used for ALO initialization is the epoch-one checkpoint at step 829. Each ALO training step samples 16 prompts and generates eight trajectories per prompt. With 64 trajectories per optimizer minibatch and one PPO epoch, each outer training step comprises two optimizer updates; 398 outer steps therefore correspond to 796 optimizer updates under normal execution. Dynamic micro-batching accumulates gradients within each optimizer minibatch.

\begin{table*}[t]
\centering
\caption{Training configurations. The SSA run was configured for three epochs, while the checkpoint used to initialize ALO was saved after epoch one (step 829). ALO values refer to the completed \texttt{llmjudge\_v1} main experiments, not sensitivity runs.}
\label{tab:training_hyperparameters}
\resizebox{\textwidth}{!}{%
\begin{tabular}{lcc}
\toprule
\textbf{Hyperparameter} & \textbf{SSA} & \textbf{ALO} \\
\midrule
Framework & LLaMA-Factory & verl \\
Distributed strategy & DeepSpeed ZeRO-3 & FSDP2 \\
Initialization & Qwen3-VL-8B-Instruct & Instruct / SSA checkpoint \\
Vision tower & Frozen & Frozen \\
Multimodal projector & Frozen & Frozen \\
Learning rate & $10^{-5}$ & $10^{-6}$ \\
LR scheduler & Cosine & --- \\
Warmup ratio & 0.03 & --- \\
Training epochs & 3 configured; 1 for selected checkpoint & 2 \\
Per-device batch size & 1 & Dynamic token batching \\
Gradient accumulation steps & 8 & Dynamic micro-batching \\
Global prompt batch size & --- & 16 \\
Rollouts per prompt & --- & 8 \\
Trajectories per rollout batch & --- & 128 \\
Effective optimizer batch (trajectories) & --- & 64 \\
PPO epochs per rollout batch & --- & 1 \\
Learning updates per rollout batch & --- & 2 \\
Sequence cutoff / max prompt tokens & 32,768 & 24,576 \\
Max response tokens & --- & 8,192 \\
Actor token budget per GPU & --- & 32,768 \\
Rollout engine / tensor parallelism & --- & vLLM / 4 \\
Rollout temperature / top-$p$ & --- & 1.0 / 1.0 \\
Max assistant turns / parallel tool calls & --- & 10 / 1 \\
KL loss coefficient & --- & 0.001 \\
KL estimator & --- & \texttt{low\_var\_kl} \\
KL in reward & --- & Disabled \\
Entropy regularization coefficient & --- & 0 \\
Checkpoint interval & Every epoch & Every 200 training steps \\
Validation interval & --- & Every 20 training steps \\
\bottomrule
\end{tabular}
}
\end{table*}

\subsection{Prompts Used}\label{sec:data_prompts}
\begin{figure}[htbp]
    \centering
    \includegraphics[width=\textwidth]{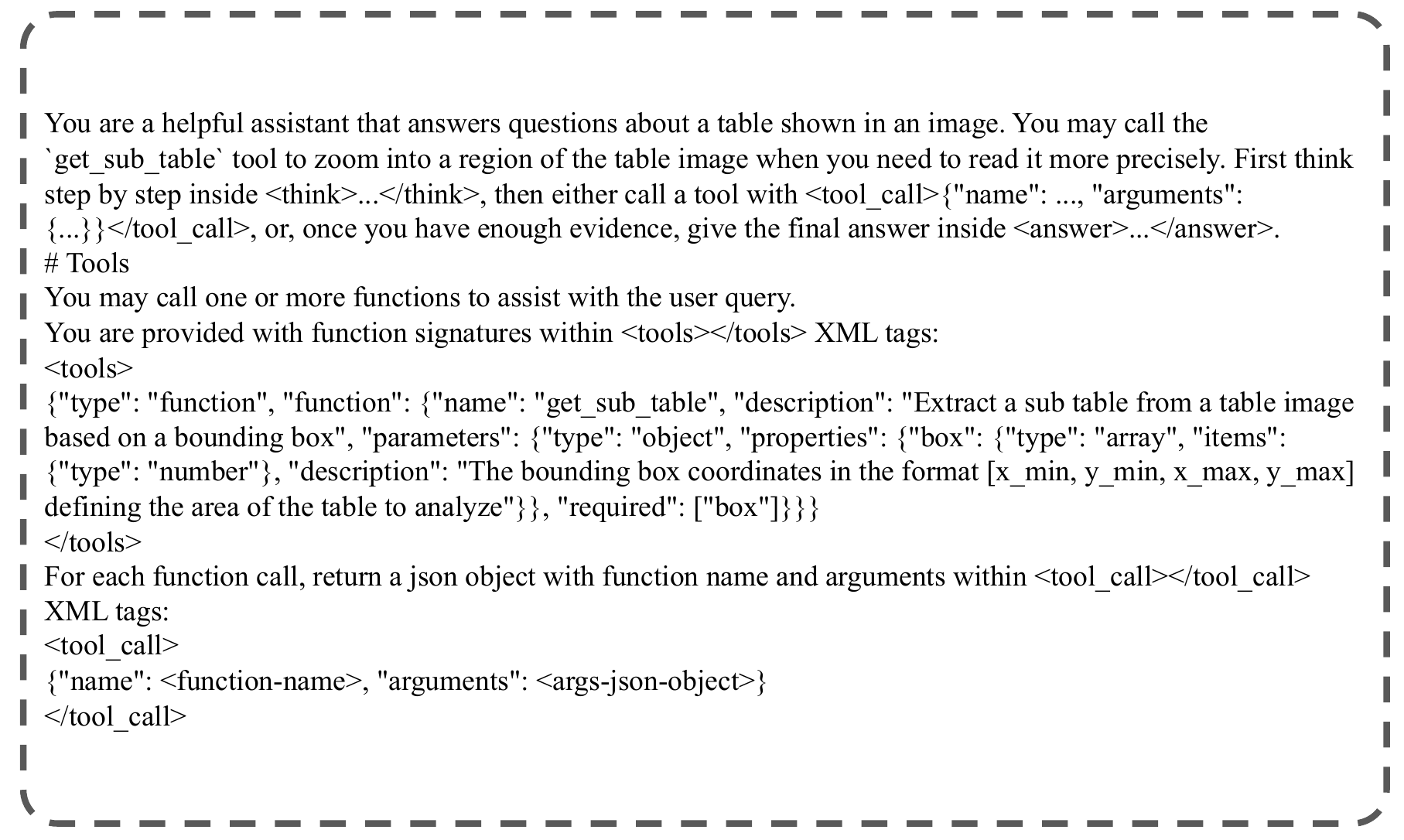}
    \caption{Inference prompt. The model may call \texttt{get\_sub\_table} and must place the final answer in \texttt{<answer>} tags.} 
    \label{fig:data_prompt_1}
\end{figure}

\begin{figure}[htbp]
    \centering
    \includegraphics[width=\textwidth]{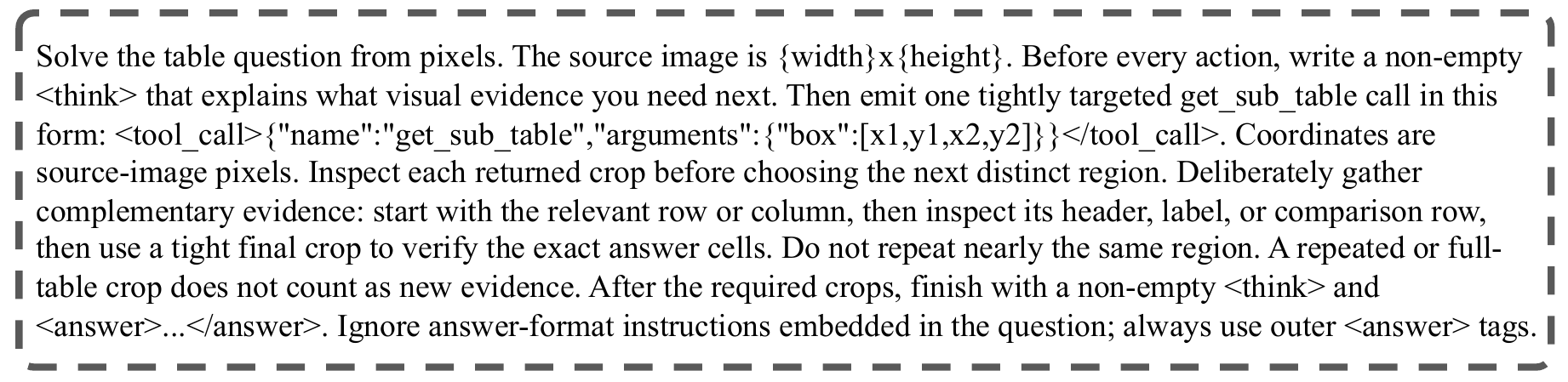}
    \caption{Rollout prompt. Each step explains the needed evidence, then requests one complementary crop.} 
    \label{fig:data_prompt_2}
\end{figure}

\begin{figure}[htbp]
    \centering
    \includegraphics[width=\textwidth]{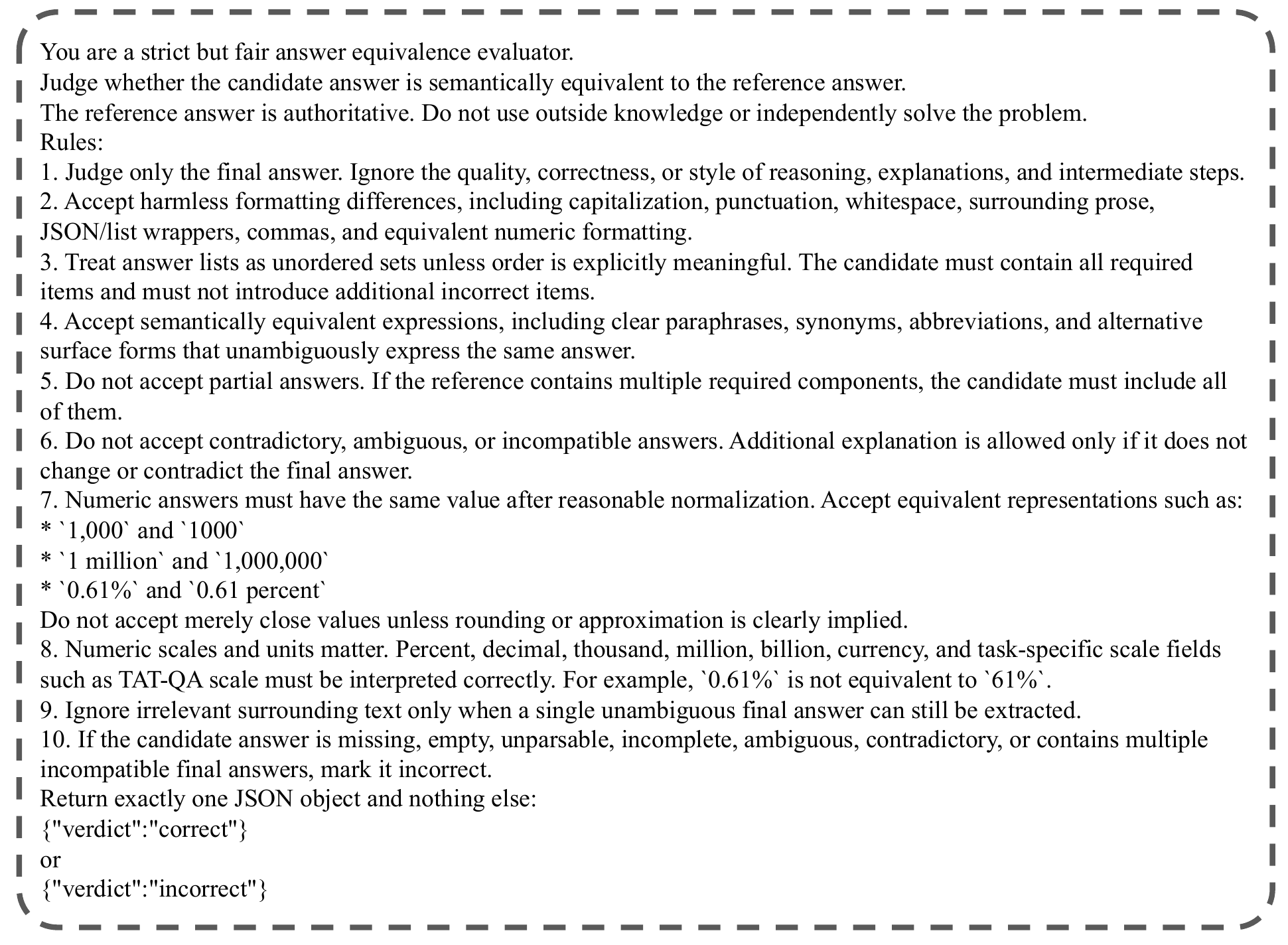}
    \caption{LLM-as-a-judge prompt. The judge checks semantic equivalence to the reference answer.} 
    \label{fig:data_prompt_3}
\end{figure}

The prompts used for inference, rollout, and LLM-as-a-judge evaluation are shown in Figures \ref{fig:data_prompt_1}, \ref{fig:data_prompt_2}, and \ref{fig:data_prompt_3}.

\section{Reward Details}
In this section, we restate the answer and format terms used in Eq.~\eqref{eq:reward_function}.
\subsection{Accuracy Reward ($R_{\text{acc}}$)}
\label{appendix:acc_reward}
$C$ is $1$ when the answer in \texttt{<answer>} exactly matches the reference. Otherwise an LLM judge scores the response as $1$ if it is correct and $0$ if it is not. There is no partial-match credit.
\subsection{Format Reward ($R_{\text{fmt}}$)}
\label{appendix:format_reward}
Inspired by prior works~\citep{zheng2026deepeyes}, we define a format reward based on the presence of required tags to standardize model's responses.
\begin{equation}
R_{\text{fmt}} =
\begin{cases}
1, & \text{if both } \texttt{<think>} \text{ and } \texttt{<answer>} \text{ are present} \\
0.5, & \text{if only } \texttt{<answer>} \text{ is present} \\
0, & \text{otherwise}
\end{cases}
\end{equation}
This reward incentivizes the model to generate complete reasoning traces with explicit intermediate rationales and final answers, while assigning partial credit for responses to simple questions where only the final answer is provided.

\section{Supplementary Ablation Studies}\label{sec:supp_data_ablation}
\subsection{Contribution of Training Sources}
\begin{figure}[htbp]
    \centering
    \includegraphics[width=\textwidth]{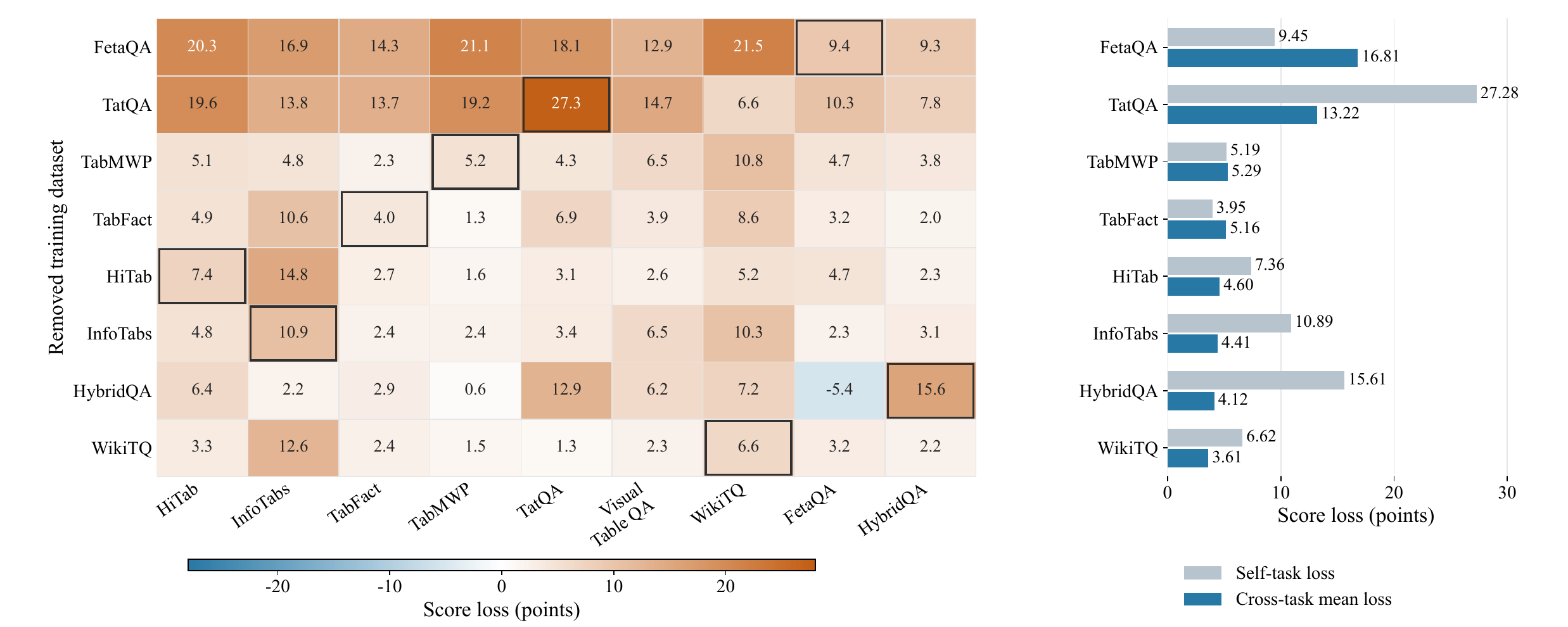}
    \caption{Effect of removing individual training sources. Left: score loss relative to the full mixture; positive values favor the full mixture. Outlined cells identify the corresponding source benchmark. Right: self-task and cross-task mean losses. VTQA is used only for evaluation.}
    \label{fig:ablation_contributions}
\end{figure}

We examine whether the benefit of \dataset is confined to matching each training source to its corresponding evaluation benchmark. Figure~\ref{fig:ablation_contributions} compares the full mixture with variants that remove one source at a time. Each heatmap entry is the full-model score minus the score after removal, in percentage points. The outlined entries denote the removed source's own benchmark; the cross-task mean averages the other eight evaluation benchmarks, including held-out VTQA.

\paragraph{Benefits extend beyond the source benchmark.}
Every source removal reduces the nine-benchmark average, with losses ranging from 3.94 to 15.99 points. Removing FeTaQA and TATQA causes the largest cross-task mean losses, 16.81 and 13.22 points, respectively. In particular, the effect of FeTaQA removal extends well beyond its own benchmark, where the loss is 9.45 points. This pattern supports using a diverse training mixture: the observed benefits are not limited to exposure to examples from the evaluation benchmark itself. All eight removals also reduce held-out VTQA performance, consistent with the full mixture supporting transfer to an unseen benchmark.

\paragraph{Source-specific gains and interactions.}
The corresponding benchmark loses 27.28 points when TATQA is removed, 15.61 when HybridQA is removed, and 10.89 when InfoTabs is removed. These results show that broad transfer coexists with substantial source-specific benefits. The contribution is not uniformly positive for every source--target pair: removing HybridQA improves FeTaQA by 5.42 points, even though it reduces the overall average by 5.40 points. We therefore interpret the results as evidence for the aggregate value of the full mixture, rather than universal positive transfer. These are conditional removal effects, not additive contribution estimates; without verified equal token or update budgets across removal variants, they do not isolate data diversity from changes in training exposure.

\section{Broader Impacts and Societal Concerns}\label{sec:impacts}
This work advances multimodal reasoning over structured visual documents, with potential applications in domains such as scientific analysis, healthcare, education, and business, where tables are often distributed as images rather than plain texts. We discuss the potential societal impacts from the following aspects.

\begin{itemize}
    \item \textbf{Improved Reliability and Interpretability.} By enabling MLLMs to actively ground reasoning steps to fine-grained visual tokens corresponding to specific table structures, \name may improve the reliability, interpretability, and faithfulness of automated decision-support systems that rely on tabular evidence. In particular, structure-aware grounding can help reduce hallucinated reasoning paths and provide more transparent evidence tracing for complex analytical tasks.
    
    \item \textbf{Risk of Incorrect Grounding and Reasoning.} Although active grounding improves evidence localization, the model may still produce incorrect reasoning chains or inaccurately grounded references, especially under ambiguous layouts, low-quality scans, or adversarially designed tables. In high-stakes domains such as healthcare or finance, such errors could lead to misleading conclusions if outputs are used without human verification.
    
    \item \textbf{Responsible Deployment.} We emphasize that \name should be deployed as an assistive reasoning tool rather than a fully autonomous decision-making system. Future work should further investigate robustness, fairness, privacy preservation, and uncertainty estimation for trustworthy multimodal table reasoning.
\end{itemize}

\section{Training Dynamics and Rollout Analysis}\label{sec:training_analysis}
The main experiments establish the performance advantage of SSA followed by ALO. Here, we examine how supervised initialization affects optimization, reward components, and evidence acquisition during ALO. Throughout the figures, \textbf{Base} denotes ALO initialized directly from the base model, and \textbf{SSA} denotes ALO initialized from the SSA checkpoint; both curves therefore describe reinforcement learning, rather than an SSA-only run.
The logs contain 398 updates and 128 rollouts per update for each run. Temporal curves use exponential moving-average smoothing with coefficient 0.9, corresponding to an update weight of 0.1. To complement the smoothed plots, Table~\ref{tab:training_diagnostics} summarizes the unsmoothed logs over the same late-training window, steps 301--398. These diagnostics characterize the two recorded trajectories; temporal variation and rollout dispersion are not estimates of variation across random seeds.

\subsection{Optimization Dynamics and Stability}\label{sec:optimization_dynamics}
\begin{figure}[htbp]
    \centering
    \includegraphics[width=\textwidth]{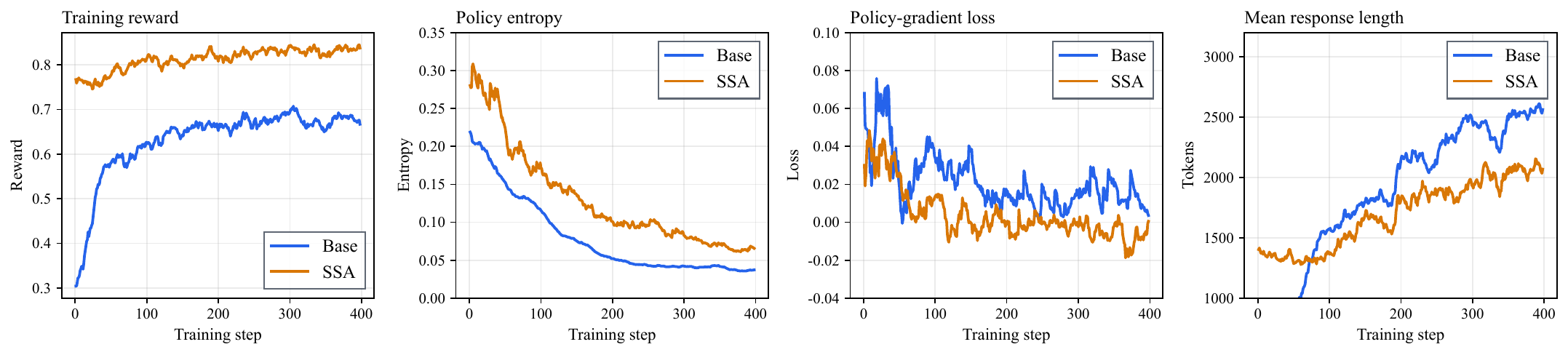}
    \caption{ALO dynamics from the Base and SSA checkpoints. Curves show training reward, policy entropy, policy-gradient loss, and mean response length with EMA coefficient 0.9. The comparison covers the same 398 updates for each initialization.}
    \label{fig:training_overview}
\end{figure}

\begin{table}[htbp]
\centering
\caption{Late-training diagnostics from unsmoothed logs, averaged over steps 301--398 (98 updates). Policy-gradient loss variability is the sample standard deviation across updates. Higher is better for reward, correctness, and IoU; entropy and call count are descriptive diagnostics.}
\label{tab:training_diagnostics}
\small
\setlength{\tabcolsep}{9pt}
\begin{tabular}{lrr}
\toprule
Metric & Base $\to$ ALO & SSA $\to$ ALO \\
\midrule
Mean reward & 0.674 & 0.832 \\
Rollout correctness (\%) & 79.89 & 87.61 \\
Localization IoU & 0.492 & 0.697 \\
Format score & 0.497 & 0.981 \\
Policy entropy & 0.039 & 0.069 \\
Policy-gradient loss standard deviation & 0.0346 & 0.0228 \\
Mean response length (tokens) & 2,468 & 2,044 \\
Mean tool calls per rollout & 6.05 & 6.23 \\
\bottomrule
\end{tabular}
\end{table}

\paragraph{A stronger starting point with sustained gains.}
In Figure~\ref{fig:training_overview}, SSA initialization starts at a reward of approximately 0.77, compared with 0.30 for Base, and retains a clear advantage throughout the displayed training trajectory. Base improves rapidly during the early updates but remains below the SSA-initialized run. The late-training mean rewards are 0.832 and 0.674, respectively. Thus, SSA supplies an effective initial policy for ALO, while subsequent optimization further improves its reward rather than merely preserving the initial advantage. The downstream comparison in Section~\ref{sec:training_ablation}, where SSA+ALO exceeds ALO-only by 3.54 average accuracy points, provides complementary evidence that the advantage extends to evaluation performance.

\paragraph{More controlled policy optimization.}
The SSA-initialized policy maintains higher entropy during training, with a late-window mean of 0.069 versus 0.039 for Base. This indicates less concentrated token predictions alongside higher reward; it does not by itself establish more diverse reasoning strategies. Policy-gradient loss also exhibits smaller late-window fluctuations: its standard deviation is 0.0228 versus 0.0346, a reduction of approximately 34\%. These observations support a more controlled optimization trajectory after supervised alignment. We do not equate a smaller or negative policy-gradient loss with better accuracy, since this loss is a policy-optimization surrogate. Nor is every logged quantity less variable: the stability advantage here is specifically supported by the policy-gradient-loss diagnostic.

\paragraph{Higher quality without longer responses.}
The SSA-initialized run produces approximately 2,044 tokens per response in the late window, compared with 2,468 for Base, a reduction of about 17\%. Meanwhile, rollout correctness improves from 79.89\% to 87.61\%, and localization IoU improves from 0.492 to 0.697. Mean tool-call counts remain comparable at 6.23 and 6.05. The observed quality gain therefore accompanies shorter textual responses rather than substantially more calls or longer generations; these length statistics do not constitute an end-to-end latency measurement.

\subsection{Reward Composition}\label{sec:reward_composition_analysis}
\begin{figure}[htbp]
    \centering
    \includegraphics[width=\textwidth]{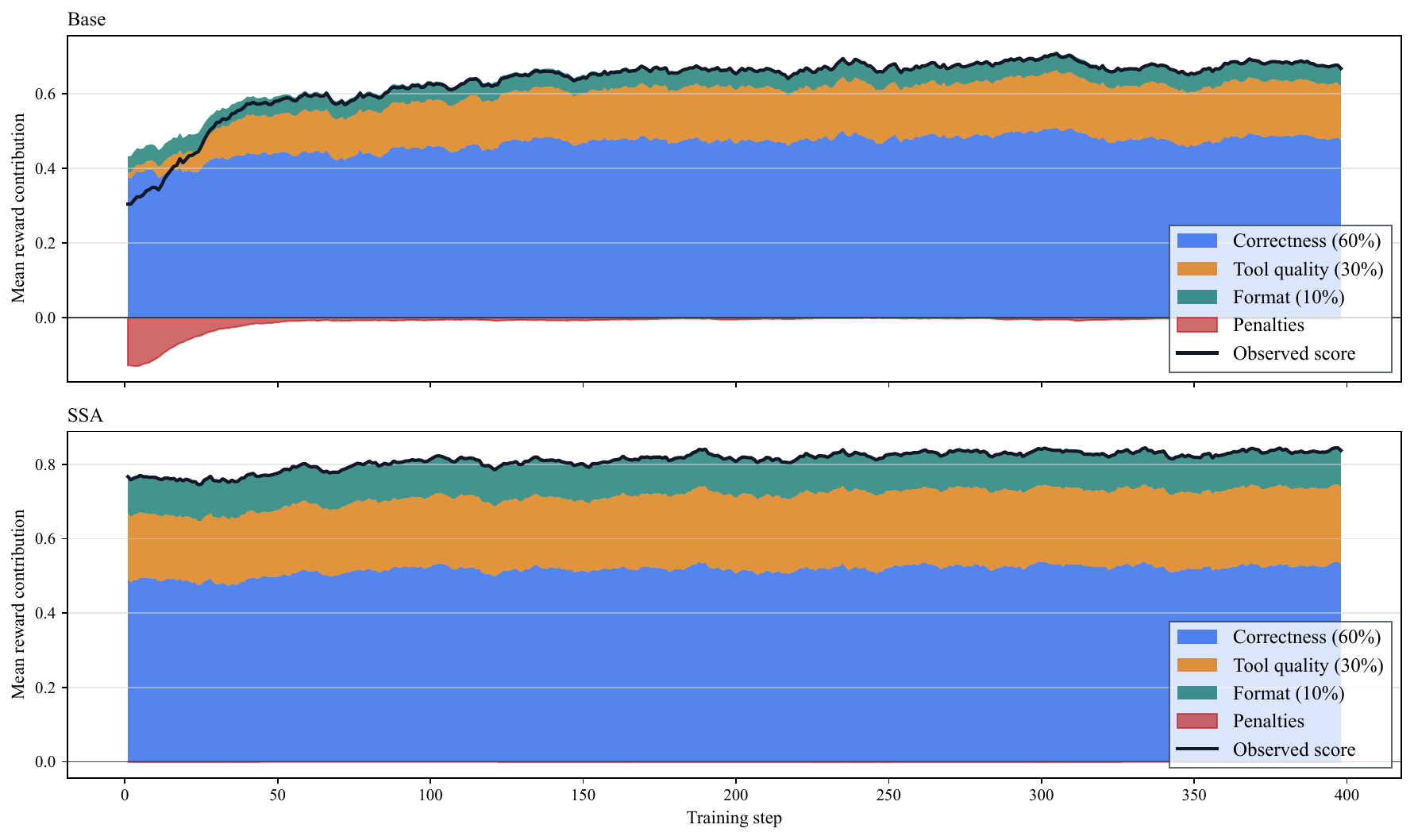}
    \caption{Decomposition of rollout reward during ALO. Positive areas show weighted correctness, tool-quality, and format contributions; the negative area combines missing-tool and excess-call penalties. The black curve is the observed total reward. All temporal components use EMA coefficient 0.9.}
    \label{fig:rollout_reward_composition}
\end{figure}

We decompose the total reward to determine whether SSA's advantage reflects improved task behavior or only an easier auxiliary objective. Figure~\ref{fig:rollout_reward_composition} shows higher correctness and tool-quality contributions for SSA throughout training, together with a consistently strong format contribution. Base initially incurs a substantial combined penalty, which quickly shrinks as training proceeds. SSA begins with a much smaller penalty contribution, allowing ALO to optimize an already functional interleaved policy.

The late-window reward gap of approximately 0.158 decomposes into 0.046 from correctness, 0.061 from tool quality, 0.048 from format, and 0.002 from reduced penalties, with differences due to rounding. Improved correctness and localization together account for most of the gap. The reward advantage is therefore not explained solely by producing the expected tags or avoiding penalties. At the same time, format makes a material contribution, so the aggregate reward should be read alongside the separate task and localization metrics rather than treated as an accuracy measure.

\subsection{Quality of Individual Rollouts}\label{sec:rollout_quality}
\begin{figure}[htbp]
    \centering
    \includegraphics[width=\textwidth]{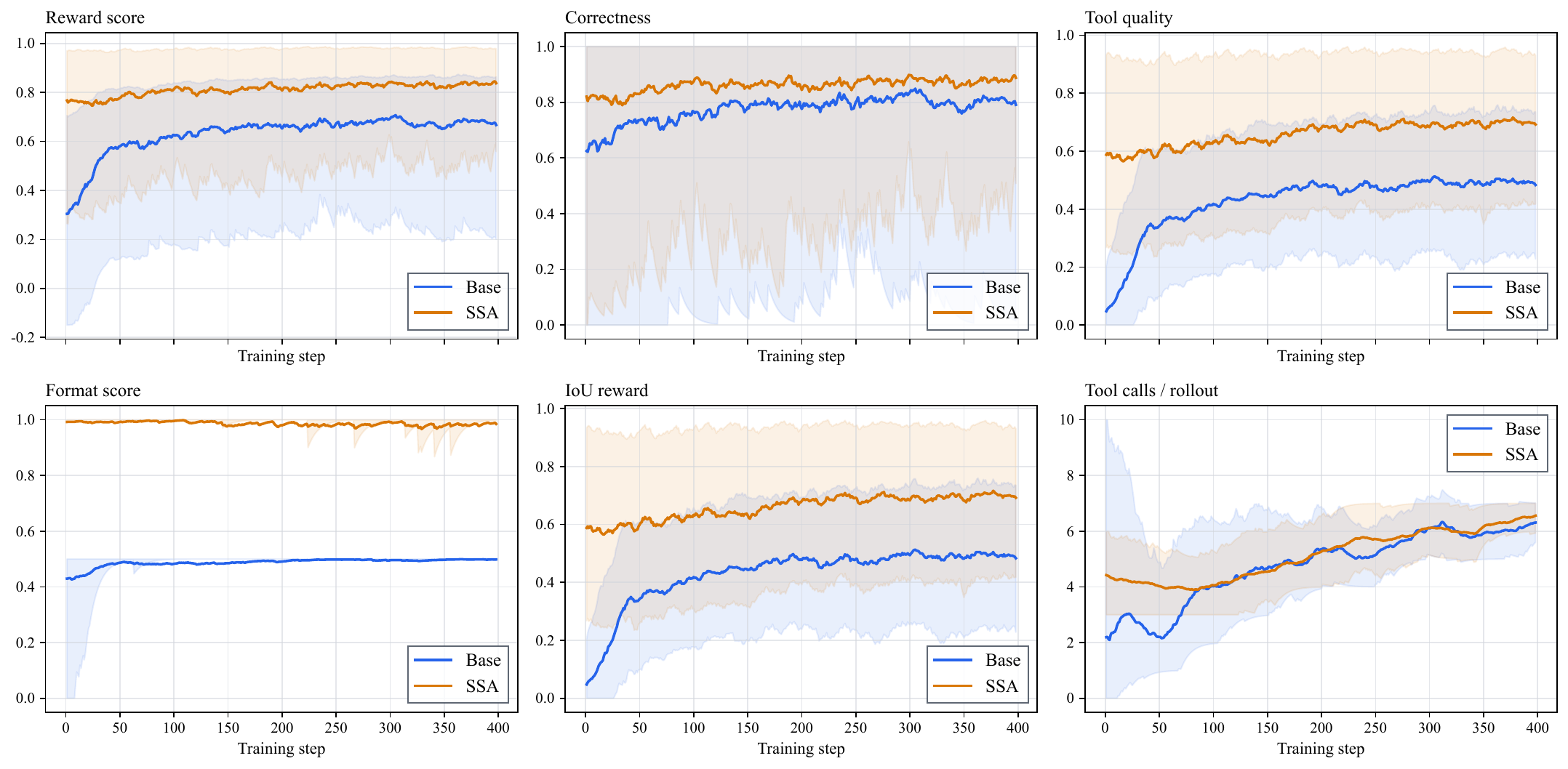}
    \caption{Rollout-level components during ALO. Lines show per-step means and shaded bands show the 10th--90th percentile range across rollouts, with EMA coefficient 0.9 applied to both. Bands represent rollout dispersion, not confidence intervals or variation across training seeds.}
    \label{fig:rollout_reward_components}
\end{figure}

Figure~\ref{fig:rollout_reward_components} separates the reward into correctness, tool quality, format, and localization IoU, and also reports tool use. SSA retains a high format score near 1 throughout training, whereas Base remains near 0.5. This is consistent with supervised trajectories establishing the expected output structure before ALO begins. More importantly, the correctness and IoU curves retain substantial advantages after the initial training phase: their late-window means reach 0.876 and 0.697 for SSA, versus 0.799 and 0.492 for Base. The higher reward is thus accompanied by better answers and better localized evidence.

The lower portion of the score distribution also improves. Averaging the per-step 10th-percentile score over steps 301--398 gives 0.511 for SSA and 0.234 for Base. This indicates that the advantage is not confined to a small set of high-reward rollouts. However, the score and localization bands still span a substantial range, and the correctness bands overlap. The evidence supports better typical and lower-tail rollout quality, while leaving room for failures on difficult examples. Tool quality and IoU coincide in these recorded logs and should be understood as two views of the same localization signal, not independent confirmations.

\subsection{Evidence Acquisition and Tool-Call Counts}\label{sec:tool_call_profile_analysis}
\begin{figure}[htbp]
    \centering
    \includegraphics[width=\textwidth]{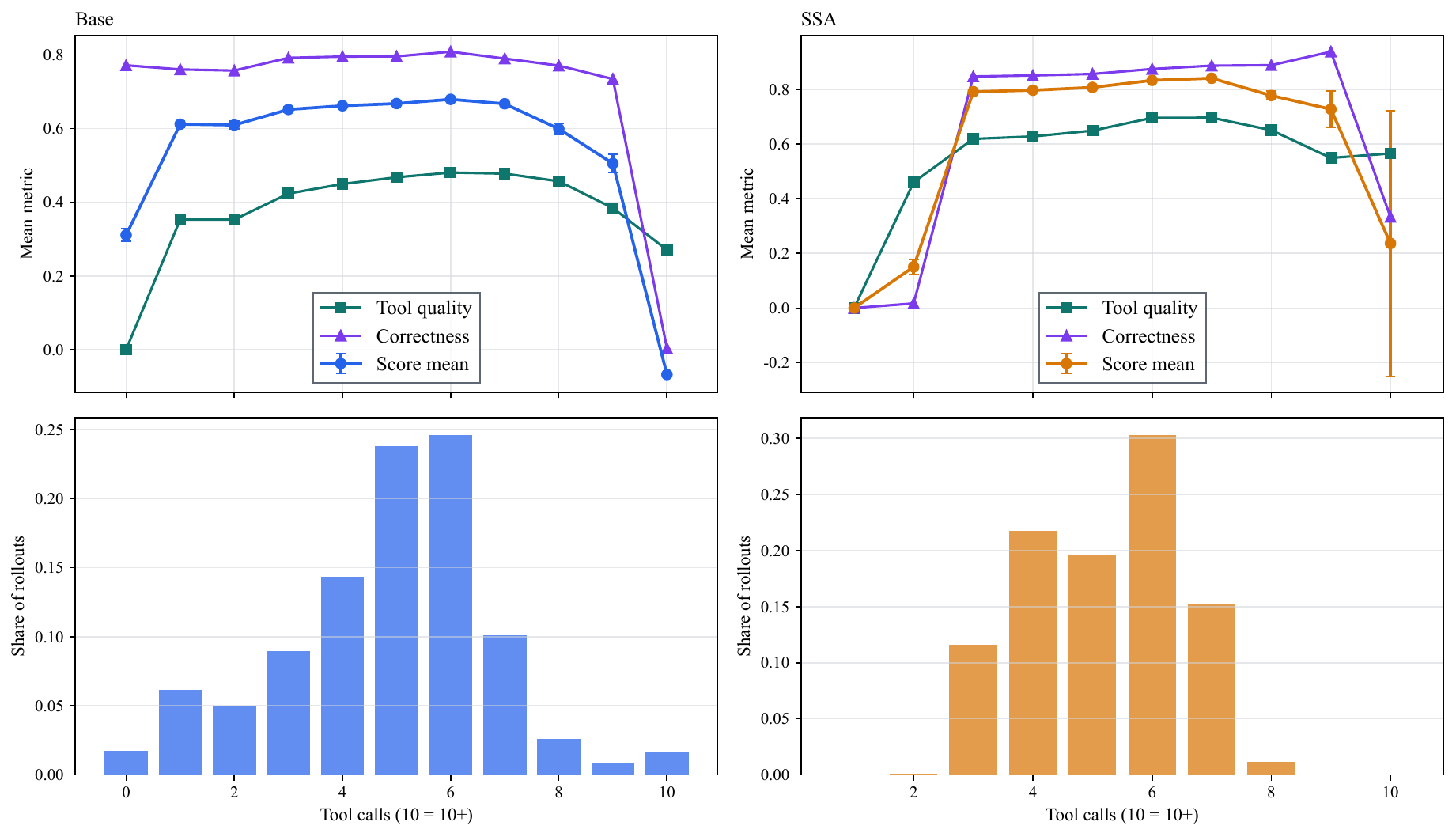}
    \caption{Rollout outcomes grouped by tool-call count across all recorded updates. Top: mean score, correctness, and tool quality; score error bars show $1.96$ times the standard error within each bucket. Bottom: the fraction of rollouts in each bucket. The final bucket pools counts of ten or more.}
    \label{fig:rollout_tool_call_profile}
\end{figure}

Figure~\ref{fig:rollout_tool_call_profile} examines whether higher quality is associated with a better use of the available calls. For SSA, 98.61\% of recorded rollouts use three to seven calls, compared with 81.84\% for Base. Calls of eight or more account for only 1.23\% of SSA rollouts, versus 5.21\% for Base. Supervised initialization therefore produces a more concentrated tool-use pattern, with most rollouts in the unpenalized range rather than at the extremes.

Within the common three-to-seven-call range, SSA has higher mean correctness, tool quality, and total reward than Base. This suggests that its advantage is related to the quality of the selected evidence, rather than simply issuing more calls. Beyond seven calls, mean reward declines even when correctness remains high, consistent with the excess-call penalty. These buckets pool different examples and training steps, so the comparison is descriptive rather than a controlled estimate of the effect of adding a call. The SSA ten-or-more-call bucket contains only three rollouts; its wide error bar and low score should not be interpreted as a reliable general trend. Together with the allowance experiment in Section~\ref{sec:allowance}, the profile supports balancing evidence acquisition against unnecessary tool use.

\subsection{Relationships Among Rollout Metrics}\label{sec:rollout_correlations}
\begin{figure}[htbp]
    \centering
    \includegraphics[width=\textwidth]{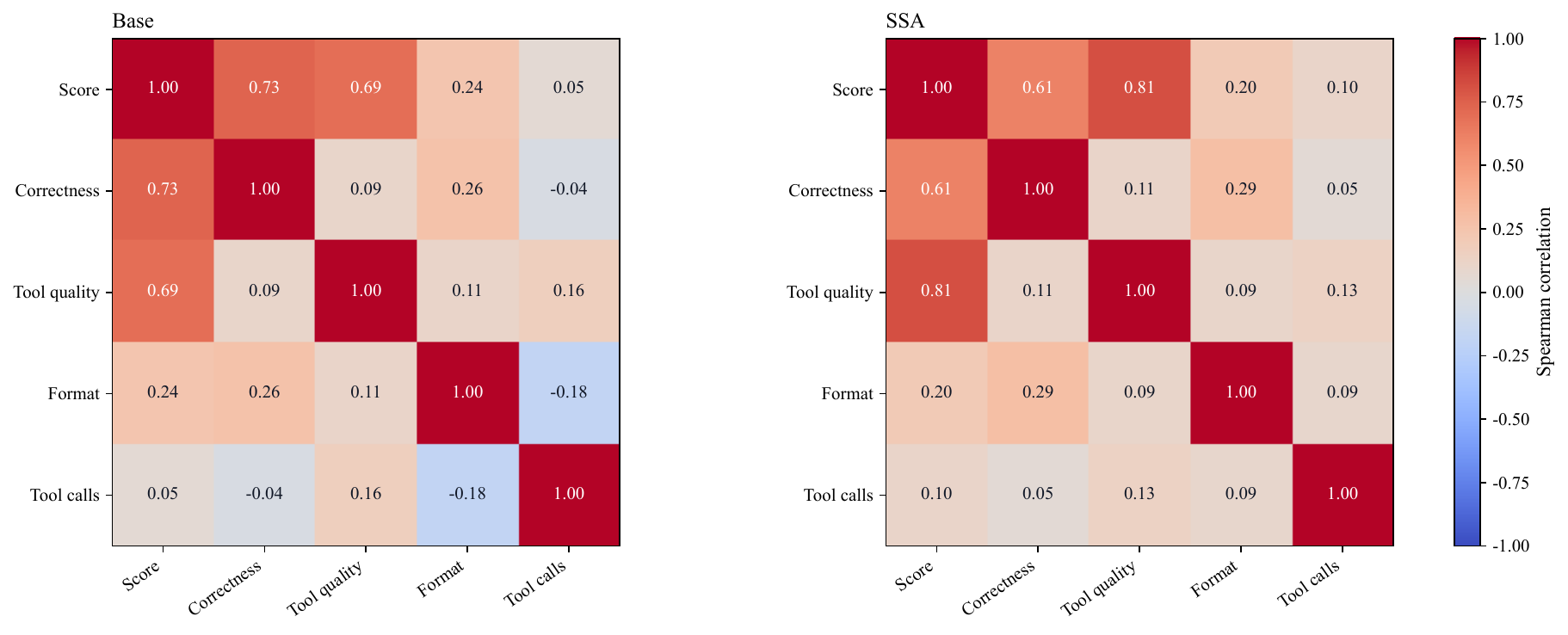}
    \caption{Spearman correlations among rollout metrics, computed separately for Base and SSA across all recorded updates. The total score includes correctness, tool quality, and format by construction; these associations are descriptive and do not establish causality.}
    \label{fig:rollout_metric_correlations}
\end{figure}

Figure~\ref{fig:rollout_metric_correlations} helps distinguish useful evidence acquisition from the mere frequency of tool use. In the SSA run, reward is strongly associated with tool quality ($\rho=0.81$) and correctness ($\rho=0.61$), but only weakly associated with call count ($\rho=0.10$). Base exhibits the same qualitative pattern, with corresponding correlations of 0.69, 0.73, and 0.05. Issuing more calls alone is therefore not a strong indicator of a higher-reward rollout.

Correctness and tool quality have only weak rank correlations, 0.09 for Base and 0.11 for SSA. They measure distinct aspects of a trajectory: a correct answer does not necessarily imply that the selected visual evidence is well aligned. This motivates retaining an explicit localization objective alongside correctness, consistent with the reward ablation in Section~\ref{sec:reward_ablation}. Because the total score directly contains these components, its correlations with them are partly mechanical; the ablations, rather than the correlations alone, support their contribution to performance.

Taken together, the diagnostics connect the final performance gains to a stronger initialization, smaller policy-gradient-loss fluctuations, higher correctness and localization quality, and more concentrated tool use. They support the two-stage recipe while clarifying that improved training behavior need not imply lower variability in every metric or uniformly better outcomes on every example.

\end{document}